\documentclass[a4paper,12pt]{article}

\usepackage{color}
\usepackage{amsfonts}
\usepackage{graphicx}
\usepackage{algorithm}
\usepackage{algorithmic}
\usepackage{enumerate}
\usepackage[hidelinks]{hyperref}
\usepackage{mathtools}
\usepackage{booktabs}
\usepackage[numbers,sort&compress]{natbib}
\usepackage{authblk}
\usepackage{subcaption}
\usepackage[table,xcdraw]{xcolor}

\algsetup{indent=2em}

\newcommand{\algword}[1]{{\ \mathbf{#1}\ }}

\title{AQMP: Image compression through Adaptive Quadtree Refinement and Matching Pursuit with Hyperparameter Optimization}

\author[1]{Franco Cerino}
\author[1]{Emmanuel Tassone}
\author[1]{Manuel Tiglio}

\affil[1]{CONICET.  
Facultad de Matemática, Astronomía, Física y Computación, Universidad Nacional de Córdoba (5000), Argentina}

\begin{document}

\maketitle

\begin{abstract}

We present AQMP, a novel image codec combining Adaptive Quadtree Refinement with Matching Pursuit. Unlike conventional Matching Pursuit methods that operate on fixed-size sub-images, AQMP dynamically adapts block sizes to local image structure, allocating finer partitions where the image is complex and coarser ones where it is smooth. This adaptivity yields superior compression ratios compared to fixed-size block Matching Pursuit at equivalent image quality, while offering significant parallelization opportunities at both the tree-leaf level and during compression of individual nodes. The algorithm is governed by user-specified accuracy and sparsity parameters alongside a small set of additional hyperparameters. To navigate the trade-off between compression efficiency and visual quality, we perform multi-objective hyperparameter optimization using the Tree-Structured Parzen Estimator, producing comprehensive Pareto fronts. Experimental results show that AQMP achieves up to $4\times$ higher compression rates than JPEG at comparable SSIM values, while maintaining competitive quality across a broad range of compression regimes. Performance evaluation is provided using a representative set of test images. To ensure reproducibility and promote adoption, we have made our implementation publicly available on GitHub under the MIT license. 
\end{abstract}
\section{Introduction}

Image compression is a fundamental aspect of digital communication and storage, enabling efficient transmission and storage of visual data. With the exponential growth of digital content, the need for effective compression techniques has become more critical than ever. By reducing the amount of data required to represent an image, these techniques save storage space and lower the bandwidth needed for transmission. A prominent example is web browsing, where faster loading times are essential, particularly on mobile devices where storage and bandwidth are often limited. More broadly, image compression plays a vital role in reducing the costs associated with data storage and transmission, making it an indispensable technology for consumers and enterprises alike.

Further applications include digital photography and medical imaging. In the latter, compression — often employing lossless or nearly lossless techniques — enables efficient storage and transmission of high-resolution images while adhering to legal and regulatory requirements that prioritize diagnostic accuracy and patient safety. This is critical for remote diagnosis and telemedicine, where preserving image fidelity is legally mandated to avoid misinterpretation \cite{Liu2017}. Image compression also plays an important role in video streaming, where compressed frames are transmitted to ensure smooth playback, although most compression in that context is performed interframe \cite{wang04, wallace91}. While modern video codecs (e.g., H.264 or HEVC) achieve significant compression efficiency through interframe techniques that exploit temporal redundancy between frames, our work focuses on intraframe compression, which instead exploits the redundancy within each individual frame.

Traditional image compression methods, such as JPEG and its successors, employ quantization to reduce the bit rate. However, this irreversibly discards high-frequency details, thereby limiting reconstruction fidelity. This paper addresses this issue by applying Matching Pursuit, a sparse approximation technique that represents the most significant structures of the image, mitigating the reliance on rigid block quantization thresholds. The objective is to achieve higher compression rates while preserving perceptual quality

The field of image compression has seen significant advances over the years. Recently, the integration of artificial intelligence (AI) and deep learning has led to notable progress in this area. Methods based on convolutional neural networks (CNNs) and recurrent neural networks (RNNs) have been employed to achieve compact image representations with high compression performance \cite{toderici17, balle17}. These AI-based methods learn from large datasets, enabling them to generalize well across diverse image types. For instance, Toderici et al.\ \cite{toderici17} proposed a recurrent neural network approach for full-resolution image compression, while Ballé et al.\ \cite{balle17} introduced an end-to-end optimized method using deep learning. Despite their impressive performance, AI-based approaches often require substantial computational resources for both training and inference, making them less suitable for resource-constrained devices.

Matching Pursuit (MP) is a family of greedy algorithms that requires no learning phase and has been widely used in image compression and compressive sensing \cite{qaisar2013cs} due to its ability to find sparse signal representations \cite{pati93}. It iteratively selects the best-matching \textit{atoms} from an over-complete dictionary to approximate a signal (as explained in detail below), achieving a practical balance between compression efficiency and image quality. The use of MP in image compression has been extensively studied, with various adaptations proposed over the years. Elad et al.\ \cite{elad94} demonstrated its effectiveness in sparse modeling of signals and images, while Nava-Tudela \cite{tudela12} explored its application in image representation and compression. MP's ability to adapt to local image structure makes it particularly suitable for images with varying levels of detail. However, traditional MP approaches typically rely on fixed-size block partitioning, which can lead to suboptimal compression ratios and image quality \cite{elad94, tudela12, mallat93}.

Prior work on MP-based image compression typically partitions the image into fixed-size $N \times N$ and operates on an over-complete dictionary of size $N^2\times M$, where $M> N^2$, seeking a sparse yet accurate representation via a greedy algorithm. 

A key limitation is that block sizes are fixed at compression time by the encoder. To overcome this, we introduce {\bf AQMP} (\textbf{A}daptive \textbf{Q}uadtree \textbf{M}atching \textbf{P}ursuit), a new codec that adaptively partitions an image into blocks of varying sizes based on local image structure, achieving better compression ratios than standard fixed-size MP at equivalent image quality.

Greedy algorithms such as those used in MP offer a different paradigm compared to AI-based methods. While the latter leverage large datasets and complex models to learn optimal compression strategies, greedy algorithms pursue iterative, step-by-step optimization to achieve sparse representations \cite{mallat93}. This makes them more interpretable and easier to implement, especially in resource-constrained settings. Greedy algorithms can also be highly efficient in real-time applications, as they require no training or large-scale data processing. That said, the two families of methods are not always directly comparable: each has its own strengths and weaknesses, and the appropriate choice depends on the specific application. AI-based methods may achieve higher compression ratios and better perceptual quality in some settings, while greedy algorithms like MP can offer faster and more resource-efficient solutions \cite{blum00, mentzer18}. Understanding these trade-offs is therefore essential when selecting a compression technique for a given use case.

A key limitation of fixed-block MP partitioning is that no single block size is universally appropriate. If blocks are too small, the encoder must store a large number of coefficient vectors, degrading compression efficiency. If blocks are too large, local image structure may not be captured accurately, resulting in poor reconstruction quality.

To address these issues, we propose AQMP, which adaptively partitions the image into blocks of varying sizes based on local structure. By recursively refining block sizes through a quadtree decomposition, the encoder captures local image content more faithfully, yielding improved compression ratios and reconstruction quality.

The remainder of this paper is organized as follows. Section~\ref{background} provides the technical background for this work, including a description of our approach in Subsection~\ref{description} and the multi-objective hyperparameter optimization (HPO) procedure in Subsection~\ref{hyperparametersearch}. Section~\ref{results} presents and discusses the experimental results. Section~\ref{discussion} concludes the paper and outlines directions for future work.

\section{Technical background} \label{background}

This section presents the technical background for this work as a linear progression, where each subsection serves as a building block that naturally leads to the next, culminating in the full description of AQMP.
Subsection~\ref{dictionaries} explains how images are represented and introduces the dictionaries used in our approach. Subsection~\ref{greedy} outlines the core idea behind greedy compression algorithms, with a focus on Matching Pursuit as a means of obtaining sparse yet accurate signal representations. Subsection~\ref{description} provides a detailed description of the AQMP method, including its implementation and operation. Finally, Subsection~\ref{hyperparametersearch} presents a multi-objective hyperparameter optimization procedure for the AQMP algorithm.

\subsection{Dictionaries and Transforms}\label{dictionaries}

Dictionaries in signal processing and image compression are collections of prototype signal elements known as \textit{atoms}, used to represent signals or images in a sparse manner. Each atom is a basic building block that can be combined linearly to approximate more complex signals. The primary goal is to achieve a sparse representation, where only a small number of atoms are needed to accurately reconstruct the original signal --- a property that is directly beneficial for compression, as it reduces the amount of data required for storage or transmission. Dictionaries can be either predefined, such as wavelet or Fourier dictionaries, or learned from data, allowing them to adapt to the specific characteristics of the signals they represent \cite{aharon06, elad10}.

Predefined dictionaries are typically based on well-established mathematical transforms such as wavelets or the discrete cosine transform (DCT), discussed in the next subsection. An alternative is to learn dictionaries directly from data using methods such as K-SVD (K-means Singular Value Decomposition) or MOD (Method of Optimal Directions) \cite{Tiglio2022}, which iteratively update the dictionary atoms to better fit the training data. When choosing a dictionary, relevant considerations include the sparsity of the resulting representation, the computational cost of any learning procedure, and the ability of the dictionary to generalize to unseen signals \cite{aharon06, engan99}.

Several transforms are commonly used to construct dictionaries for signal and image representation. The wavelet transform is widely used due to its ability to capture both spatial and frequency information, making it well-suited for signals with localized features. The discrete cosine transform (DCT) is another popular choice, particularly in image compression standards such as JPEG, owing to its computational efficiency and its tendency to concentrate signal energy into a small number of coefficients for smooth signals. The Fourier transform is also commonly employed, especially for signals with periodic or stationary components, as it provides a global frequency decomposition using complex exponential basis functions. Unlike the DCT, which produces real-valued coefficients and is well-suited for decorrelation in compression, the Fourier transform captures both magnitude and phase information across the entire signal, making it more appropriate for spectral analysis of periodic phenomena \cite{mallat99, strang99, aharon06}.

For a 1D DCT of length $N$, the DCT matrix $C$ is an $N \times N$ matrix whose elements $C_{k,n}$ are defined as
\begin{equation}
	C_{k,n} = \alpha_k \cos\left[\frac{\pi}{N} \left(n + \frac{1}{2}\right) k \right],
\end{equation}
where $k$ and $n$ range from $0$ to $N-1$, and $\alpha_k$ is a normalization factor given by
\begin{equation}
	\alpha_k = \begin{cases} 
	  \sqrt{\frac{1}{N}} & \text{if } k = 0 \\
	  \sqrt{\frac{2}{N}} & \text{if } k > 0.
	\end{cases}
\end{equation}

\begin{figure}[h!]
\centering
\includegraphics[width=\linewidth]{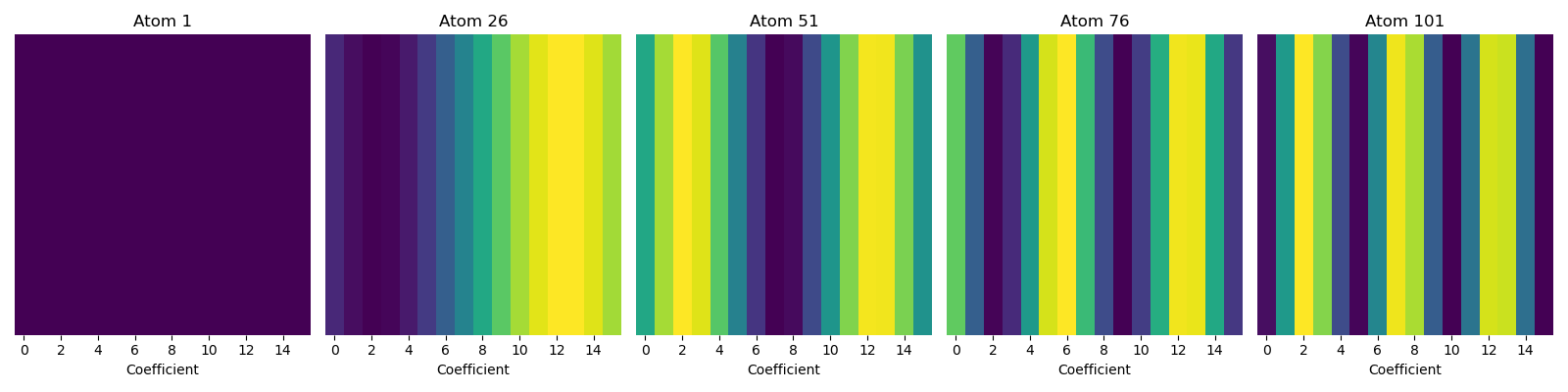}
\caption{DCT matrix transforms. The image shows matrix transforms of size $16\times128$. Each matrix is multiplied by the image to be compressed, activating different columns. Blue colors indicate columns with zero values, yellow indicates columns with unit values, and intermediate colors correspond to values between zero and one. The more yellow the columns, the greater the weight given to the corresponding image region.}
\label{fig:DCT-transform}
\end{figure}

For a 2D DCT, which is the variant used in this work, the transformation is applied separately along the rows and columns of the image, and the full 2D DCT matrix is constructed as the outer product of two 1D DCT matrices. Figure~\ref{fig:DCT-transform} illustrates the structure of these 2D DCT matrices, whose entries represent the cosine basis functions used to decompose an image into its frequency components.

As shown in Fig.~\ref{fig:DCT-transform}, different 2D DCT matrices --- commonly referred to as atoms --- activate different spatial regions of the image. For example, when atom 26 is applied to an image, only the right-hand portion is activated while the left-hand portion is suppressed. These atoms are grouped into a dictionary that plays a central role in the AQMP algorithm, described in detail in Section~\ref{description}.

\subsection{Greedy Algorithms and Matching Pursuit}\label{greedy}

Greedy algorithms \cite{knuth1980algorithms,taocp1,taocp2,taocp3,taocp4a,taocp4b} have been widely used in image and video compression due to their simplicity and efficiency. These algorithms iteratively make locally optimal choices in pursuit of a global optimum. Notable examples include the Set Partitioning in Hierarchical Trees (SPIHT) algorithm, which has been extensively used for wavelet-based image compression \cite{said96}, and the Embedded Zerotree Wavelet (EZW) algorithm, which also employs a greedy approach to achieve efficient image compression \cite{shapiro93}. Both have significantly advanced the field by providing effective methods that balance compression ratio and image quality \cite{Kim1997}.

The greedy algorithm used in this work is Matching Pursuit (MP), a technique that iteratively selects the best-matching atoms from a dictionary to build a sparse approximation of a signal. Let $\text{MP}(\mathbf{D}, \mathbf{y}, \varepsilon, k)$ denote an MP algorithm whose goal is to solve Problem~(\ref{MP_problem},\ref{stopping_criteria}):

\begin{equation}\label{MP_problem}
	\min_{\mathbf{x}}\left \| \mathbf{x} \right \|_0 
	\;\; \textup{subject to} \;\; 
	\left \| \mathbf{Dx - y} \right \|_2 < \varepsilon 
\end{equation}

where $\mathbf{D}$ is a dictionary, $\mathbf{y}$ is the target signal, $\varepsilon$ is the accuracy tolerance, and $k$ is the maximum number of iterations. The stopping condition is
\begin{equation}\label{stopping_criteria}
	\left \| \mathbf{Dx - y} \right \|_2 < \varepsilon
	\;\; \vee \;\; 
	\left \| \mathbf{x} \right \|_0 = k \, . 
\end{equation}

The rationale for this stopping criterion is as follows. MP initializes $\mathbf{x}$ as the zero vector. At each greedy iteration, one previously zero component of $\mathbf{x}$ is assigned a non-zero value, so that after $k$ iterations $\mathbf{x}$ has exactly $k$ non-zero entries. Without the sparsity constraint on $k$, the accuracy condition $\|\mathbf{Dx - y}\|_2 < \varepsilon$ could in principle be satisfied with a large number of non-zero coefficients, yielding a representation that is insufficiently sparse and therefore achieves little compression.

The MP algorithm proceeds as follows:

\begin{enumerate}
	\item Given a signal \( x \in \mathbb{R}^N \) and an over-complete dictionary \( D = \{ \phi_i \}_{i=1}^M \) where \( \phi_i \in \mathbb{R}^N \), find a sparse representation of \( x \) as a linear combination of atoms from $D$.

	\item
	\begin{enumerate}
	   \item Initialize the residual \( r^0 = x \) and the iteration counter \( k = 0 \).
	   \item At each iteration \( k \), select the atom \( \phi_{i_k} \) that best matches the current residual:
	   \[
	   i_k = \arg \max_i \left| \langle r^k, \phi_i \rangle \right|,
	   \]
	   where $\langle \cdot, \cdot \rangle$ denotes a scalar product defined on $\mathbb{R}^N$.
	   \item Update the residual:
	   \[
	   r^{k+1} = r^k - \langle r^k, \phi_{i_k} \rangle \phi_{i_k}.
	   \]
	   \item Increment \( k \) and repeat steps (b)--(c) until the stopping criterion is met.
	\end{enumerate}
\end{enumerate}

In this work, each image (or image block) is treated as a flattened vector of pixel values. Specifically, an ($N \times N$) block is rearranged into a single column vector ($\mathbf{x} \in \mathbb{R}^{N^2}$), making the 2D image data amenable to Matching Pursuit. We employ Orthogonal Matching Pursuit (OMP) rather than standard MP, as it offers better convergence properties \cite{pati93}; specifically, we use the \texttt{OrthogonalMatchingPursuit} class from \texttt{scikit-learn}.

The main advantage of MP lies in its ability to adapt to the local structure of images, providing a balance between compression efficiency and image quality. For images with complex local structures, more iterations are performed, adding more atoms to the sparse representation and achieving a more faithful approximation of the original image. The parameters involved in MP include the dictionary size, the stopping criterion, and the number of iterations. MP has been shown to be effective in various applications, including image compression \cite{elad94, pati93}, sparse signal modeling \cite{mallat93}, and image representation \cite{tudela12}. Despite its effectiveness, traditional MP methods often rely on fixed-size block partitioning, which can lead to suboptimal compression ratios and image quality \cite{elad94, tudela12, mallat93}.

\subsection{Adaptive Quadtree Refinement and Matching Pursuit (AQMP)} \label{description}

The AQMP method introduced in this paper addresses the limitations of fixed-block partitioning by adaptively partitioning the image into blocks of varying sizes based on local image structure. Adaptivity is achieved through a quadtree structure, where each block is recursively divided into four quadrants if the target accuracy and sparseness are not met. This process continues until the desired accuracy and sparseness are achieved or the minimum block size is reached.

The encoder seeks a sparse representation of sub-images of size $N\times N$, where $N$ is location-dependent. The process begins with the user specifying an accuracy tolerance $\varepsilon$ in the $\ell_2$ norm, a sparseness parameter $s$ equal to the number of MP iterations $k$ and the $\ell_0$ norm of the sparse representation, and the minimum and maximum block sizes $N_{min}$ and $N_{max}$. Setting a minimum block size is motivated by the fact that very small blocks contain little information to compress, while a maximum block size limits computational cost. By design, $N_{max}$ and $N_{min}$ in AQMP are related by a power of $2$: $N_{min} = L$ and $N_{max} = 2^i L$, for integers $L$ and $i$. This defines a set of admissible block sizes
$$
S = \{ L, 2L, \dots, 2^i L \}
$$
for some $L > 0$ and $i \geq 0$. 

A collection of dictionaries $\mathbf{D}_N$ of size $N \times M$, one for each $N \in S$, is assumed to be available. The construction of these dictionaries is not the focus of this paper, and any suitable dictionary may be used, though some choices may be more appropriate depending on the image type and domain. A popular choice, such as that discussed in Sec.~\ref{dictionaries}, is to concatenate a DCT (Discrete Cosine Transform) basis with wavelets (e.g., Haar \cite{Tiglio2022}). Other extensions, such as adding Jacobi polynomials \cite{Tiglio2022}, are also supported.

A sparse representation is then sought using MP. If the representation does not meet the $\ell_2$ target within $k$ iterations for a given sub-image, that sub-image is divided into four quadrant blocks of size $N/2 \times N/2$, and a sparse representation of each block is sought in turn. This partitioning is repeated for any block that fails to meet the target $\varepsilon$ within $k$ iterations. The refinement continues until all blocks satisfy both the accuracy and sparseness requirements, or the minimum block size is reached. In summary, AQMP strikes a delicate balance between accuracy and compression that fixed-size block MP cannot achieve, as the latter offers little flexibility to control this trade-off.

The AQMP method offers several advantages over traditional fixed-block approaches. By adaptively partitioning the image according to local structure, AQMP achieves better compression ratios while maintaining high image quality. This is particularly beneficial for images with varying levels of detail, where fixed-block methods may fail to capture local structure accurately. Additionally, the quadtree structure combined with the efficiency of the Matching Pursuit algorithm makes AQMP suitable for real-time applications (cf.\ Sec.~\ref{description}).

In contrast to other quadtree-based partitioning techniques that are often tailored to specific image types or tasks \cite{Gross1996,Shaffer86,Samet1984}, AQMP is designed to compress images of any kind and resolution. For example, Gross et al.\ \cite{Gross1996} employ wavelet transforms to decompose surface data into different frequency components; while their algorithm targets low complexity, the need for an inverse wavelet transform at each triangulation step can be computationally intensive for large datasets. By employing a flexible quadtree structure with adaptive refinement, AQMP ensures that both high-detail and low-detail regions are optimally segmented, making it versatile and effective across diverse imaging scenarios.

The following subsections describe the two components of the image codec introduced here: the encoder and the decoder.

\begin{figure}[h!]
    \centering
    \includegraphics[width=\linewidth]{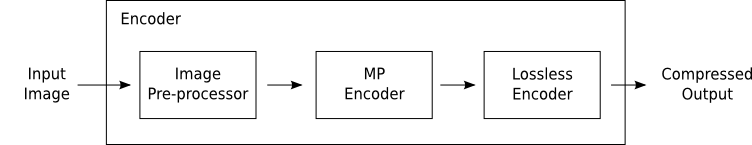}
    \caption{Encoder high level design.}
    \label{fig:encoder}
\end{figure}

\begin{figure}[h!]
    \centering
    \includegraphics[width=\linewidth]{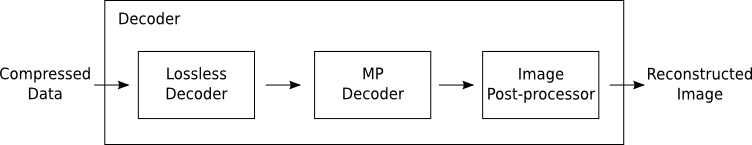}
    \caption{Decoder high level design.}
    \label{fig:decoder}
\end{figure}

\subsubsection{Encoder} \label{description-encoder}

At a high level, the encoder consists of three steps: (i) image pre-processing, (ii) Adaptive Quadtree Refinement with MP compression, and (iii) lossless coding. See Figure~\ref{fig:encoder} for a graphical overview.

Steps (i) and (iii) are lossless; Step (ii) is lossy. Although a full codec is presented here, the core contribution of this work is Step (ii), which is described in detail following the high-level description below.

\begin{enumerate}[(i)]
\item The input image is pre-processed to serve as input to Step (ii). The color channels are converted to one of the supported formats, such as RGB or $\mathrm{YC_BC_R}$, and each channel is then processed independently.

\item Each channel is divided into blocks of size $N \times N$. A Matching Pursuit algorithm seeks a sparse representation of each block. If MP cannot find a solution meeting the target accuracy and sparseness for a given parent block, that block is split into four children quadrants of size $N/2 \times N/2$ (assuming $N$ to be even). Compression is then applied to each child block at the next level. This process is repeated on each branch until the target accuracy and sparseness are achieved everywhere in the image.

\item A lossless coding step is applied to the output of Step (ii) to exploit the redundancy of the sparse vectors. The implementation uses DEFLATE \cite{deutsch96}, though any lossless algorithm may be substituted. The output of this step is the final bitstream.
\end{enumerate}

Given the list of color channels produced by Step (i), AQMP is defined by the following two algorithms, presented as pseudocode.

\begin{algorithm}
\caption{$mp\_encoder$}
\label{mp-coder}
\begin{algorithmic}
\REQUIRE $channels, \varepsilon$
\ENSURE $X$
\STATE $X := [\;]$
\FORALL{$c \algword{in} channels$} 
	\STATE $X.append(mp\_encoder\_rec(c, \varepsilon, N_{max}))$
\ENDFOR
\end{algorithmic}
\end{algorithm}

\begin{algorithm}
\caption{$mp\_encoder\_rec$}
\label{mp-coder-rec}
\begin{algorithmic}
\REQUIRE $channel, \varepsilon, N$
\ENSURE $X$
\STATE $k := min\_sparsity(\varepsilon, N)$
\STATE $blocks := $ list of blocks of size $N\times N$ from $channel$
\STATE $X := [\;]$
\FORALL{$b \algword{in} blocks$} 
	\STATE $y := unravel(b)$
	\STATE $x := MP(\mathbf{D}_N, y, \varepsilon, k)$
	\IF{$\left \| x \right \|_0 \geq  k$ \AND  $N > N_{min}$}
		\STATE $X.concat(mp\_encoder\_rec(b, \varepsilon, N / 2))$
	\ELSE
		\STATE $X.append(x)$
	\ENDIF
\ENDFOR
\end{algorithmic}
\end{algorithm}

\begin{figure}[h!]
    \centering
\makebox[\textwidth][c]{%
    \includegraphics[width=\paperwidth]{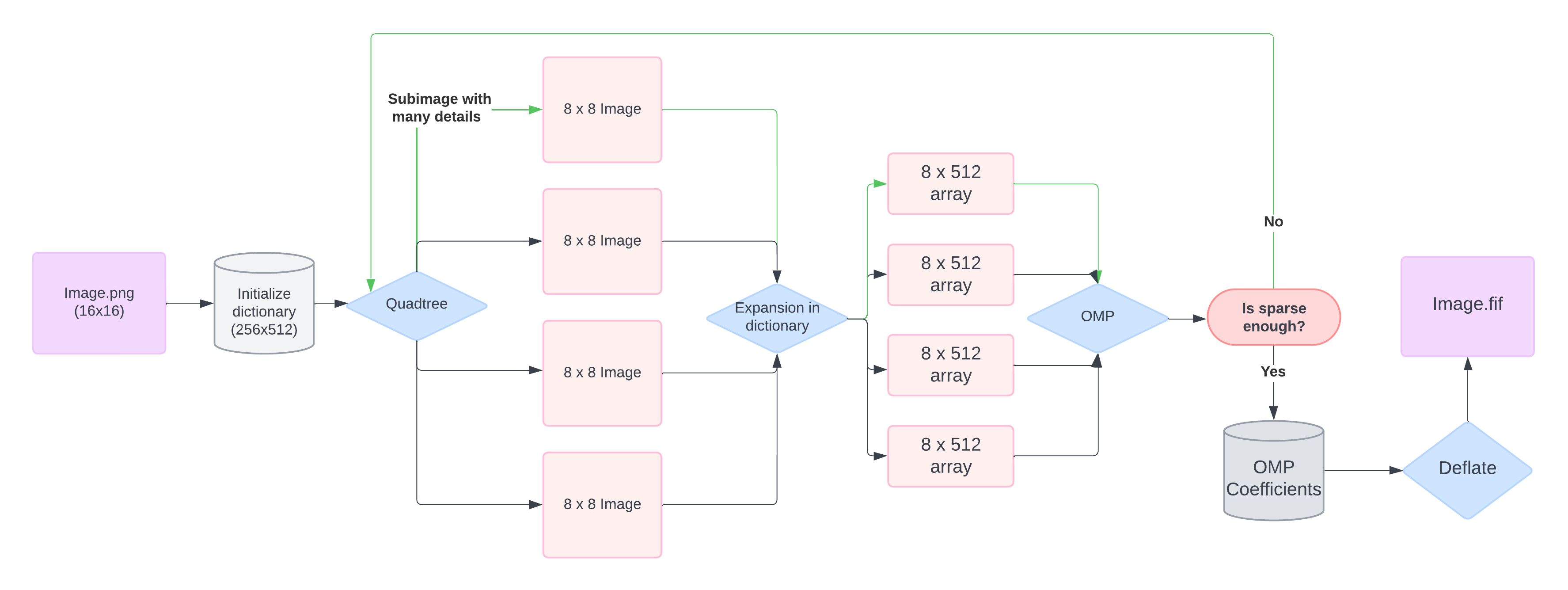}
} 
    \caption{AQMP-compression path flow. }
    \label{fig:code-flow}
\end{figure}

In these pseudocodes, $unravel$ flattens a two-dimensional block into a one-dimensional vector, while $min\_sparsity$ returns the minimum expected sparsity for a given block size $N$ and error tolerance $\varepsilon$, determining whether further refinement with a smaller block is warranted. The approach is not tied to any particular implementation of these functions. A visualization of the encoder logic is shown in Fig.~\ref{fig:code-flow}, illustrating the compression path for an image of size $16 \times 16$ pixels.

\subsubsection{Decoder} \label{description-decoder}

The decoder is also composed of three steps, each reverting one step of the encoder; see Figure~\ref{fig:decoder}.

\begin{enumerate}[(i)]
\item The lossless encoding is reversed, producing a bitstream of MP-coded data. This bitstream consists of a list of encoded channels, each of which is in turn a list of sparse vectors.
\item Each encoded channel is decoded using the $mp\_decoder$ algorithm described below.
\item If necessary, a post-processing step converts the channels back to RGB, the standard format used by displays.
\end{enumerate}

Given the width $w$ and height $h$ of the original image and a list of sparse vectors $X$ for a given channel, the decoder reconstructs the channel through the following algorithm:

\begin{algorithm}
\caption{$mp\_decoder$}
\label{mp-decoder}
\begin{algorithmic}
\REQUIRE $X, w, h$
\ENSURE $channel$
\STATE $channel := Matrix(h, w) $
\FORALL{$x \algword{in} X$} 
	\STATE $N := size(x)$
	\STATE $y := \mathbf{D}_N \cdot x$
	\STATE $b := unravel^{-1}(y)$	
	\STATE $set\_block(b, channel)$
\ENDFOR
\end{algorithmic}
\end{algorithm}

The $set\_block(b, channel)$ function places block $b$ into the matrix $channel$ at the position determined by the quadtree traversal order.

\subsection{Hyperparameter Optimization (HPO)} \label{hyperparametersearch}

To perform AQMP, different parameters must first be defined before, which determine how the process is carried out, affecting both the final image quality and compression. 
The set of hyperparameters to take into account in our AQMP approach are the following:
\begin{itemize}
    \item  \texttt{max\_error}: maximum squared norm of the residual (\texttt{tol} in the utilized OrthogonalMatchingPursuit class of \texttt{scikit-learn}). 
    \item \texttt{min\_sparsity}: the threshold for the minimum allowable sparsity (or density of non-zero coefficients). This parameter works alongside \texttt{max\_error} as stopping criteria for the optimization problem. AQMP aims to solve a least squares problem with a given tolerance (\texttt{max\_error}) while using at most a maximum number of atoms (\texttt{min\_sparcity}).
	\item \texttt{min\_n}: the minimum block size.
    \item \texttt{max\_n}: the maximum block size.
   \item  \texttt{a\_cols}: the number of columns of the dictionary. 
\end{itemize}

It is convenient use an efficient methodology to search for a good combination of these parameters. A natural approach is to use an hyperparameter optimization strategy of the field of Machine Learning (ML).

Hyperparameters are configuration parameters of a ML model that must be set before training begins: unlike ML parameters, hyperparameters  do not change during training. Each combination of hyperparameters results in a unique model configuration, directly affecting its performance. To evaluate the impact of each hyperparameter combination, it is necessary to train a model for each configuration.

Standard methods, such as random and grid searches have several drawbacks, such as: i) being computationally inefficient, as they do not use the results of previous evaluations to select better models; if the hyperparameter space is very large or if the models are expensive to train, an exhaustive search is computationally expensive (but they are parallelizable). For those reasons in this work we used Sequential Model-Based Optimization
\cite{bergstra11, ozaki20} to search a good combination of hyperparameters for AQMP. 

\subsection{Sequential Model-Based Optimization (SMBO)}

SMBO is a sequential method where, in each iteration:
\begin{enumerate}
\item An ML model is trained with hyperparameters \( x \), and a cost function \( f(x) \) is evaluated.
\item The pairs \( (x, f(x)) \) are used to estimate the hyperparameters \( x \) that minimize \( f(x) \).
\end{enumerate}

In Python, the Optuna library \cite{Cerino2023} provides an efficient SMBO-based hyperparameter optimization using the Tree-Structured Parzen Estimator (TPE) approach. TPE supports continuous, discrete, and categorical hyperparameters. Optuna also supports multi-objective optimization.

\subsection{Tree-Structured Parzen Estimator (TPE)}
\begin{figure}[h!]
\centering
\includegraphics[width=0.8\linewidth]{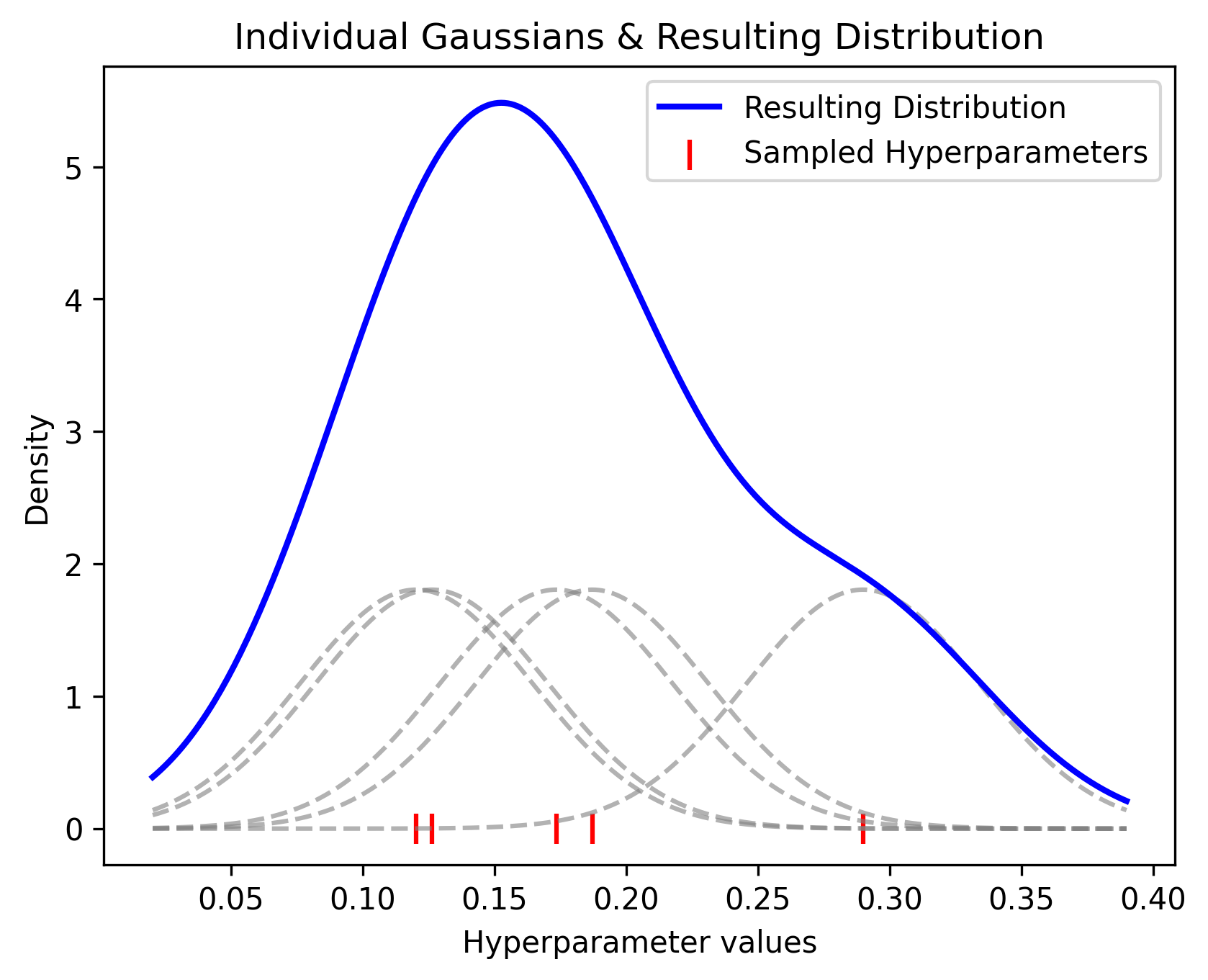}
\caption{Illustrative probability distribution built from a set of hyperparameters. Each hyperparameter taken in account is represented with a red tick and a gaussian is centered around each one to build the resulting distribution. With TPE two of these are built: one for the best hyperparameters and another for the worst. }
\label{fig:prob_dist_gaussians}
\end{figure}

\begin{itemize}
    \item In each iteration, hyperparameters are ordered based on performance on the cost function \( f(x) \), resulting in two distributions:
    \begin{itemize}
        \item \( l(x) \) for the best hyperparameters.
        \item \( g(x) \) for the worst hyperparameters.
    \end{itemize}
    \item Each distribution is created using a Gaussian kernel, resulting in a normalized sum of Gaussians (see an example in Figure \ref{fig:prob_dist_gaussians}).
\end{itemize}

\begin{figure}[h!]
\centering
\includegraphics[width=0.8\linewidth]{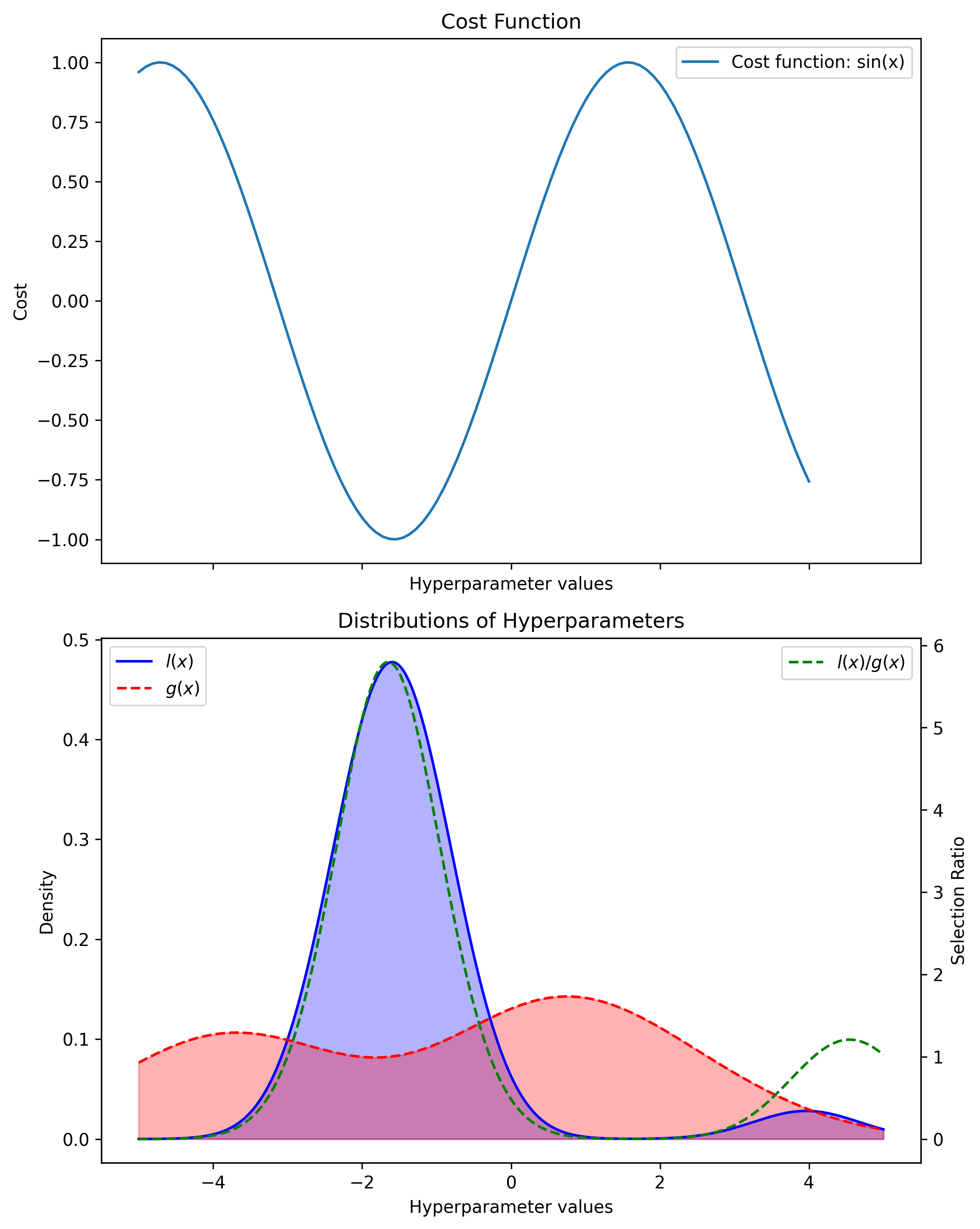}
\caption{Illustrative cost function and associated distributions of best and worst hyperparameters ($l(x)$ and $g(x)$, respectively). Also the ratio $l(x)/g(x)$ is graphed. The distributions are obtained after 50 steps with TPE Sampler.}
\label{fig:cost_function_and_hyps}
\end{figure}

\begin{itemize}
    \item The hyperparameter combination \( x \) selected for training in the next iteration is the one that maximizes the ratio \( l(x)/g(x) \) (see Figure \ref{fig:cost_function_and_hyps}).
	\item The computational cost of TPE in Optuna is \( \mathcal{O}(d n \log n) \), where \( d \) is the dimensionality of the search space and \( n \) is the number of trials~\cite{optunasamplers}. On the other hand, the cost of the random sampler is \( \mathcal{O}(d) \).
\end{itemize}

\subsubsection{Expected Improvement (EI) Criteria}

The EI criteria is a theoretical justification of why TPE chooses the hyperparameter that minimizes $\l(x)/g(x)$ to improve the performance of a model. As a summary, the main points are (for more details, see \cite{Cerino2023}):

\begin{itemize}
    \item The cost function \( f(x) = y \) is the objective to be optimized. For the purpose of selecting an approach, we choose to maximize it, noting that both approaches of maximizing or minimizing are equivalent by shifting the cost function.
	
    \item The goal is for the next hyperparameter $x^*$ from combinations \( x \) to satisfy:
    \begin{equation}
    f(x^*) - f(x) > 0
    \end{equation}
    \item We define an improvement function:
    \begin{equation}
    \text{I}(x) := \max(f(x^*) - f(x), 0)
    \end{equation}
    \item Since \( f(x) \) is unknown, it is treated as a random variable; the goal is to find \( x^* \) that maximizes the expected improvement:
    \begin{equation}
    \text{EI}(x) := \mathbb{E}[\max(f(x^*) - f(x), 0)] \, , 
    \end{equation}
\end{itemize}
where the Expected Improvement can be formulated as:
\begin{align}
\text{EI}(x) &= \int_{-\infty}^{\infty} \max(y^* - y, 0) \, p(y \mid x) \, dy \\
&= \int_{-\infty}^{y^*} (y^* - y) \, \frac{p(x \mid y) \, p(y)}{p(x)} \, dy 
\end{align}
with each the integrals are evaluated using:
\begin{equation}
p(x \mid y) = 
\begin{cases} 
l(x)  & \text{if } y < y^* \\
g(x)  & \text{if } y \geq y^* \, .
\end{cases}
\end{equation}
Here $y^*$ is the $\gamma$ quantile of the observed values, i.e. $p(y < y^*) = \gamma$. This leads to:
\begin{equation}
\text{EI}(x) \propto \left( \gamma + (1 - \gamma) \frac{g(x)}{l(x)} \right)^{-1} \, . 
\end{equation}
In summary, the goal is to find an $x^*$ that maximizes the ratio $l(x) / g(x)$.

\section{Results} \label{results}

In this section, we measure the performance of the AQMP algorithm by evaluating the trade-offs between compression rate and image quality. We define the compression rate CR as 
\begin{equation}
	\text{CR} = \frac{\text{size}_{\text{original}}}{\text{size}_{\text{compressed}}},
\end{equation}
where each size is calculated as the number of bytes in the corresponding file. For $size_{compressed}$, this means calculating the size of the reconstructed image from the compressed file.

To measure the image quality we employed the Structural Similarity Index Measure (SSIM), which is defined as
\begin{equation}
	\text{SSIM}(x, y) = \frac{(2\mu_x\mu_y + C_1)(2\sigma_{xy} + C_2)}{(\mu_x^2 + \mu_y^2 + C_1)(\sigma_x^2 + \sigma_y^2 + C_2)},
\end{equation}

where \(x\) denotes the original image and \(y\) the compressed image. Here, \(\mu_x\) and \(\mu_y\) are the mean intensities of \(x\) and \(y\), \(\sigma_x\) and \(\sigma_y\) are their respective variances, and \(\sigma_{xy}\) is the covariance between them. The constants \(C_1\) and \(C_2\) are small positive values (often defined as \(C_1 = (K_1 L)^2\), \(C_2 = (K_2 L)^2\) with \(K_1\) and \(K_2\) as calibration parameters and \(L\) the dynamic range of the pixel intensities) used to maintain numerical stability when denominators are near zero. 

SSIM is a metric which quantifies the similarity between the original and compressed images, it ranges from 0 to 1, with higher values indicating greater similarity. We use SSIM to evaluate the visual quality of the compressed images, comparing them to the original ones. 

\begin{figure}[h!]
	\centering
	\makebox[\textwidth][c]{%
	    \includegraphics[width=0.2\paperwidth]{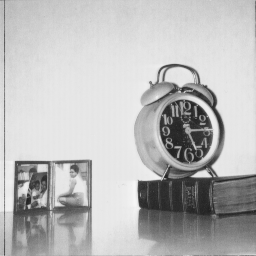}
	    \includegraphics[width=0.2\paperwidth]{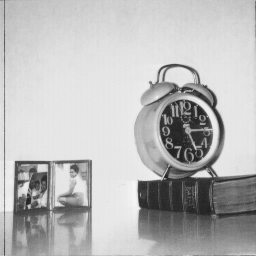}
	    \includegraphics[width=0.2\paperwidth]{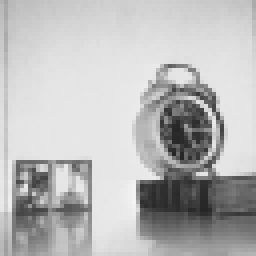}
		\includegraphics[width=0.2\paperwidth]{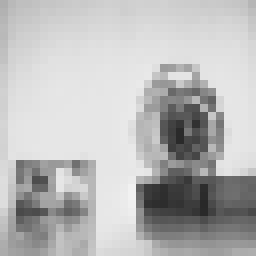}
	}
	    \caption{Image compression results for the Clock.png image with resolution $256\times256$ pixels. The image was taken from the USC-SIPI Image Database \cite{sipi}. From left to right: original image, compressed images with SSIM 0.993 and CR 1.356, SSIM 0.798 and CR 5.059, SSIM 0.653 and 11.288.}
		\label{fig:compression_results}
\end{figure}

As a proof of concept, Fig. [\ref{fig:compression_results}] shows the original image and three compressed counterparts, with a high value of SSIM and a high compression ratio  respectively. We can qualitatively see that the compressed image with high SSIM is visually indistinguishable from the original image, while the compressed image with highest compression ratio has barely visible objects. This conceptually demonstrates the usual trade-off between image quality and compression rate. The AQMP approach can produce a highly compressed image with high compression ratio ideal for scenarios where only a rough outline of the image is needed, such as movement detection task or geographic maps.

On the other hand, each compression in Fig. [\ref{fig:compression_results}] was done with a specific set of hyperparameters (we discuss what those hyperparameters are below) that are not trivial to find a priori. Finding optimal or nearly optimal hyperparameters to achieve the best trade-off between high compression and high quality is not a trivial task in our AQMP approach (in fact, in machine learning in general). We discuss and tackle this problem with hyperparameter optimization in the next section.

\subsection{AQMP Hyperparameter Optimization}\label{aqmp-hyperparameter-optimization}

We present the results of hyperparameter optimization with the TPE Sampler approach, as introduced in Section [\ref{hyperparametersearch}], using the open source package Optuna \cite{Tiglio2022}, which aims to find optimal hyperparameter configurations in optimization, in this case for AQMP to achieve a balance between the tradeoff of image quality and compression. Our analyses focus on the Pareto front, i.e. the set of all solutions where no single objective (SSIM or compression rate) can be improved without worsening the other \cite{Tiglio2022}. In this case, the pareto front illustrates the trade-offs between compression rate and SSIM for test images with varying resolutions, selected from the USC-SIPI Image Database \cite{sipi}.
This database is a collection of digitized images, which is maintained primarily to support research in image processing, image analysis, and machine vision. Images in the database can be found in mainly three sizes: 256x256 pixels, 512x512 pixels, or 1024x1024 pixels. 
We also provide a summary of the optimal parameters found for the AQMP algorithm. A detailed explanation of how the hyperparameter optimization was performed is provided in Sec. [\ref{hyperparametersearch}].

The optimization procedure focuses on balancing two key objectives: maximizing the Structural Similarity Index Measure (SSIM) and achieving an optimal compression ratio. To address this {\it multi-objective optimization problem}, we employ a Pareto optimization, which efficiently navigates the high-dimensional parameter space to identify trade-offs between SSIM and compression ratio using a Bayesian approach. 

\begin{figure}[h!]
	\centering
	\makebox[\textwidth][c]{%
	    \includegraphics[width=0.6\paperwidth]{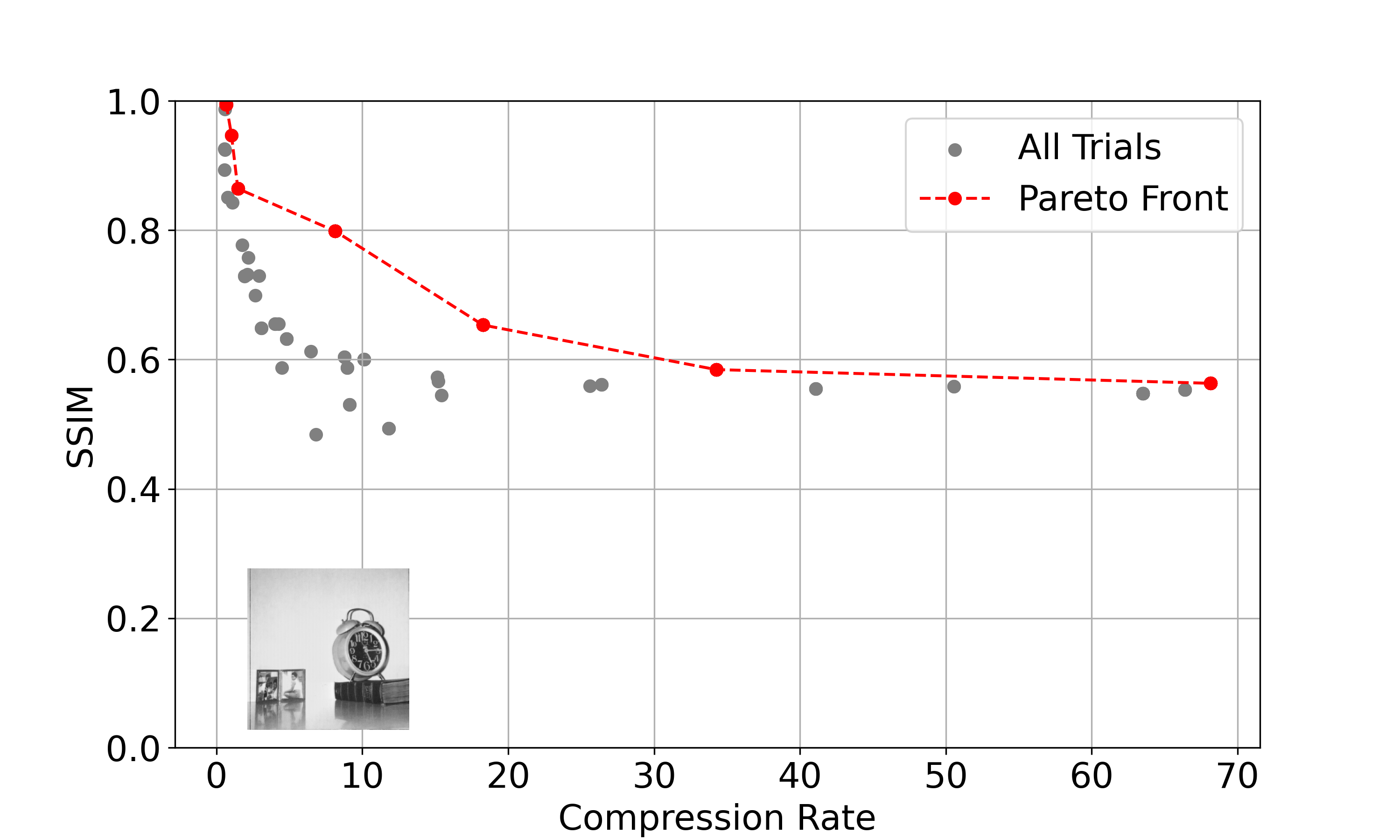} 
	}
	    \caption{Hyperparameter Optimization for Clock image (256x256). The image shows the tradeoff between the SSIM and compression ratio for the five hyperparameters. The red curve shows the Pareto front in each image, which establishes the optimal election for the tradeoff between SSIM and compression rate.}
		\label{fig:pareto_curve1}
\end{figure}

\begin{figure}[h!]
	\centering
	\makebox[\textwidth][c]{%
	    \includegraphics[width=0.6\paperwidth]{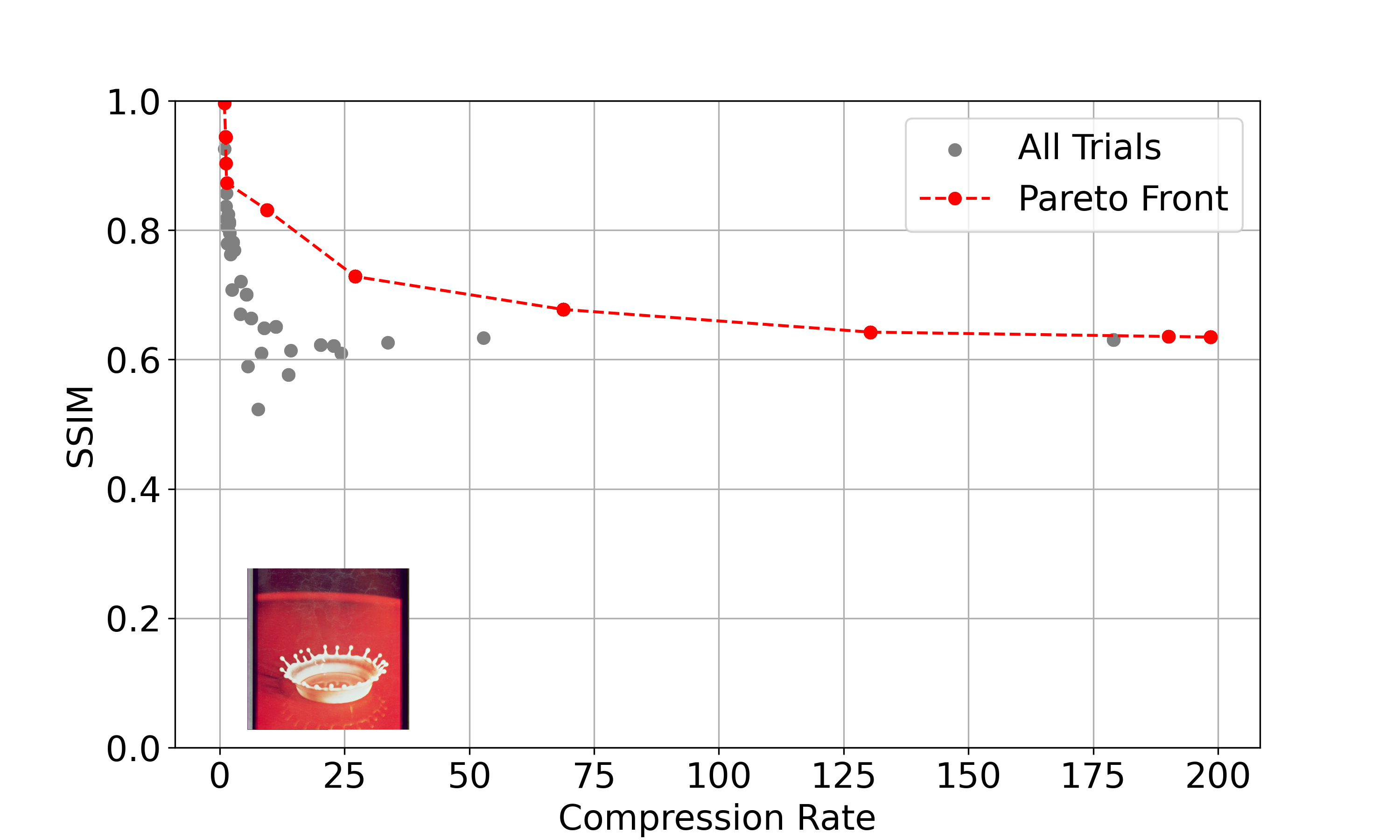} 
	}
	    \caption{Hyperparameter Optimization for Splash image (512x512). The image shows the tradeoff between the SSIM and compression ratio for the five hyperparameters. The red curve shows the Pareto front in each image, which establishes the optimal election for the tradeoff between SSIM and compression rate.}
		\label{fig:pareto_curve2}
\end{figure}

\begin{figure}[h!]
	\centering
	\makebox[\textwidth][c]{%
	    \includegraphics[width=0.6\paperwidth]{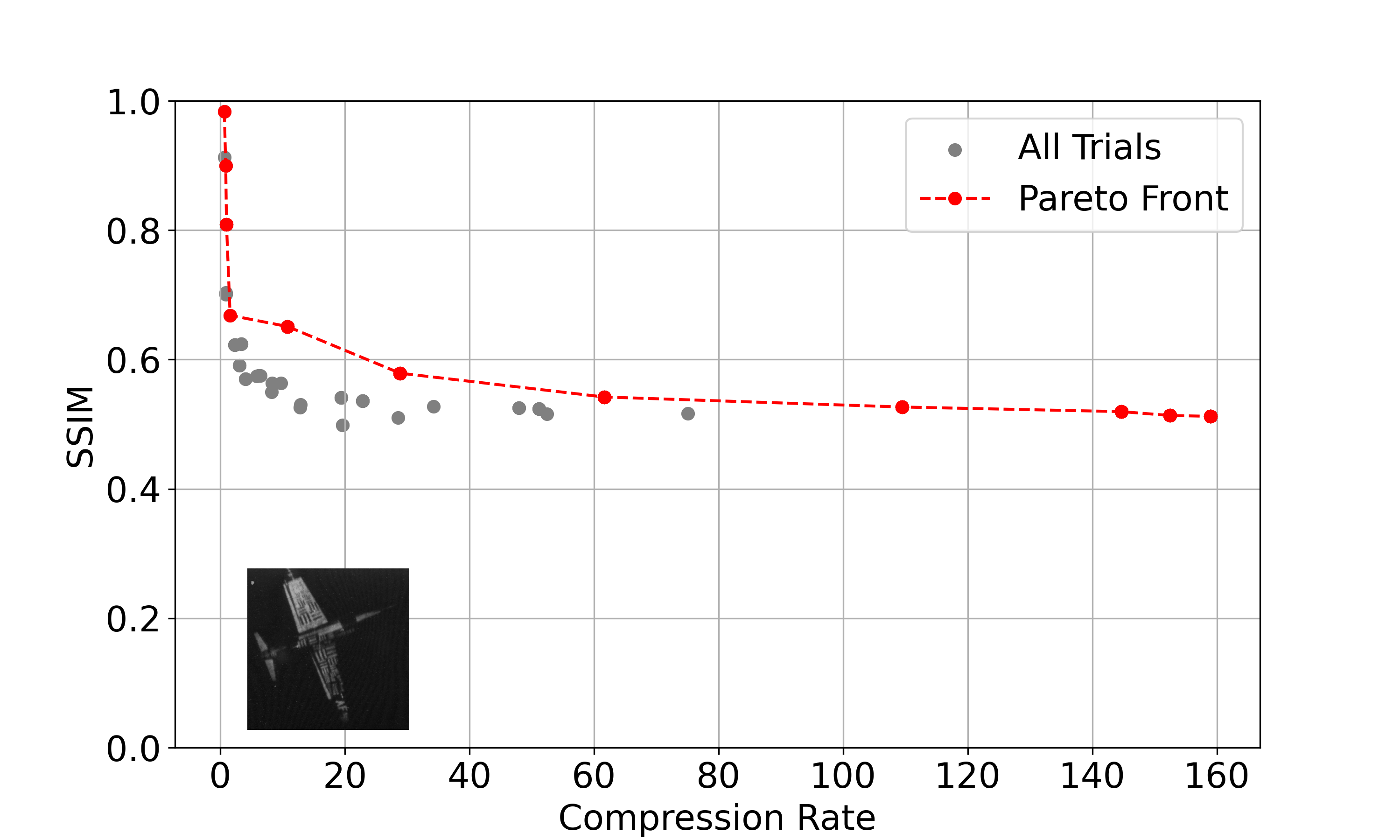} 
	}
	    \caption{Hyperparameter Optimization for Airplane (U-2) iamge (1024x1024). The image shows the tradeoff between the SSIM and compression ratio for the five hyperparameters. The red curve shows the Pareto front in each image, which establishes the optimal election for the tradeoff between SSIM and compression rate.}
		\label{fig:pareto_curve3}
\end{figure}

The plots in Figs. \ref{fig:pareto_curve1}, \ref{fig:pareto_curve2} and \ref{fig:pareto_curve3} depict the Pareto front for AQMP applied to images with three different resolutions:   Clock (256x256), Splash (512x512) and Airplane (U-2) (1024x1024). These images were taken from  the USC-SIPI Image Database \cite{sipi}. The horizontal and vertical axes show the compression rates and the SSIMs, respectively. The gray dots represent all trials conducted during the parameter optimization processes for the three images considered, illustrating the trade-offs between compression rates and SSIMs. The red dots and the dashed lines show the Pareto front, which visualizes the best trade-offs. These curves serve as valuable tools for selecting parameters that meet specific application requirements, such as prioritizing high SSIM for quality-critical applications or achieving high compression ratios where storage or transmission constraints are the driving factor(s).

At low compression rates (close to 0), the SSIM remains nearly 1, reflecting high-quality reconstructions with negligible loss of detail. And, as expected, as the compression rate increases, the SSIM begins to decline, indicating that more aggressive compression introduces visible artifacts or quality loss. In the intermediate region, the Pareto front exhibits a gradual decline in SSIM with increasing compression rate, showcasing the algorithm's ability to balance between size reduction and visual quality. At higher compression rates (above 30), the SSIM stabilizes or declines more slowly, suggesting diminishing returns, as further compression significantly affects quality without proportional gains in size reduction. 

One aspect to notice is that the compression rate is below 1 when the SSIM is very high. This occurs because, in the highly compressed file, metadata (such as file headers, compression information, or auxiliary data) can take up a significant portion of the total file size. This overhead becomes prominent when the actual compression of image content is minimal, similar to the behavior observed in formats like TIFF \cite{Tiglio2022}.This aspect becomes particularly significant in low-resolution images, such as $256\times 256$, where the overall file size is relatively small. Consequently, the proportion of metadata within the compressed file becomes more relevant, as it represents a larger fraction of the total file size.

From this plot, a few conclusions can be drawn. First, the algorithm is highly effective in preserving quality at low compression rates, making it suitable for applications requiring high fidelity. Second, as the compression rate increases, the trade-off becomes more apparent, and the Pareto front helps identify optimal parameter configurations depending on whether compression efficiency or image quality is prioritized. Finally, the presence of metadata-driven compression inefficiencies at low rates suggests that the algorithm could be further optimized by reducing unnecessary metadata or streamlining file structure, which would improve its performance for high-quality reconstructions.

Table \ref{tab:splash_table} summarizes optimal parameters found by AQMP. The results show the hyperparameter values for \texttt{min\_sparsity}, \texttt{max\_error}, \texttt{min\_n},  \texttt{max\_n} and \texttt{a\_cols}, along with the corresponding compression rate and SSIM for the Splash image. 
The total number of trials used were 100 and the computing time was of 16:57 minutes with a Intel Core i9-9980HK (8 cores).
Those hyperparameters listed in Table \ref{tab:splash_table} are located in the Pareto front, highlighting the effective balances achievable through hyperparameter optimization. This table provides a practical reference for settings that yield optimal compression quality and efficiency for this particular image at different resolutions.

\begin{table}[h!]
	\centering
	\resizebox{\textwidth}{!}{
	\begin{tabular}{cccccccc}
		\toprule
		SSIM & Compression Rate & Max Error & Min Sparsity & min\_n & max\_n & a\_cols & Execution time (s) \\
		\midrule
		0.9962 & 0.9172 & 0.0015 & 0.2084 & 4 & 4 & 256 & 65.5338 \\
		0.9032 & 1.2377 & 0.0082 & 0.3298 & 8 & 8 & 32 & 16.9183 \\
		0.8314 & 9.4504 & 0.0016 & 0.6256 & 4 & 4 & 2 & 28.6854 \\
		0.8314 & 9.4504 & 47.2277 & 0.4394 & 4 & 4 & 2 & 30.7702 \\
		0.7284 & 27.1236 & 261.8908 & 0.9236 & 4 & 8 & 2 & 7.4893 \\
		0.7284 & 27.1236 & 424.6535 & 0.8930 & 4 & 8 & 2 & 7.8326 \\
		0.6774 & 68.8056 & 0.0022 & 0.7279 & 8 & 16 & 2 & 2.4761 \\
		0.6774 & 68.8056 & 0.0032 & 0.6841 & 8 & 16 & 2 & 2.8326 \\
		0.6424 & 130.3471 & 0.0055 & 0.2468 & 8 & 32 & 2 & 1.0816 \\
		0.6356 & 190.1261 & 30.2875 & 0.2801 & 8 & 128 & 4 & 0.6998 \\
		0.6347 & 198.5337 & 0.0373 & 0.5549 & 64 & 128 & 2 & 0.7511 \\
		0.6347 & 198.5337 & 48.6796 & 0.6815 & 8 & 128 & 2 & 0.7679 \\
		\bottomrule
		\end{tabular}
	}
	\caption{Compression results for Splash image. Each row has results that are in the Pareto front. Execution time is the time needed to compress and decompress an image.}
	\label{tab:splash_table}
	\end{table}

\subsection{AQMP vs other compression algorithms} \label{AQMP-comparison}

In this subsection, we compare the compression performance different alternatives: 
\begin{itemize}
	\item AQMP with Tree-Structured Parzen Estimator (TPE) for hyperparameter optimization.
	\item AQMP with a random sampler for hyperparameter optimization.
	\item Matching Pursuit (MP) with standard fixed blocks of 8x8 and without quadtree refinement.
	\item JPEG algorithm from the Python library PIL.
\end{itemize}

\begin{figure}[h!]
    \centering
    \includegraphics[width=0.8\linewidth]{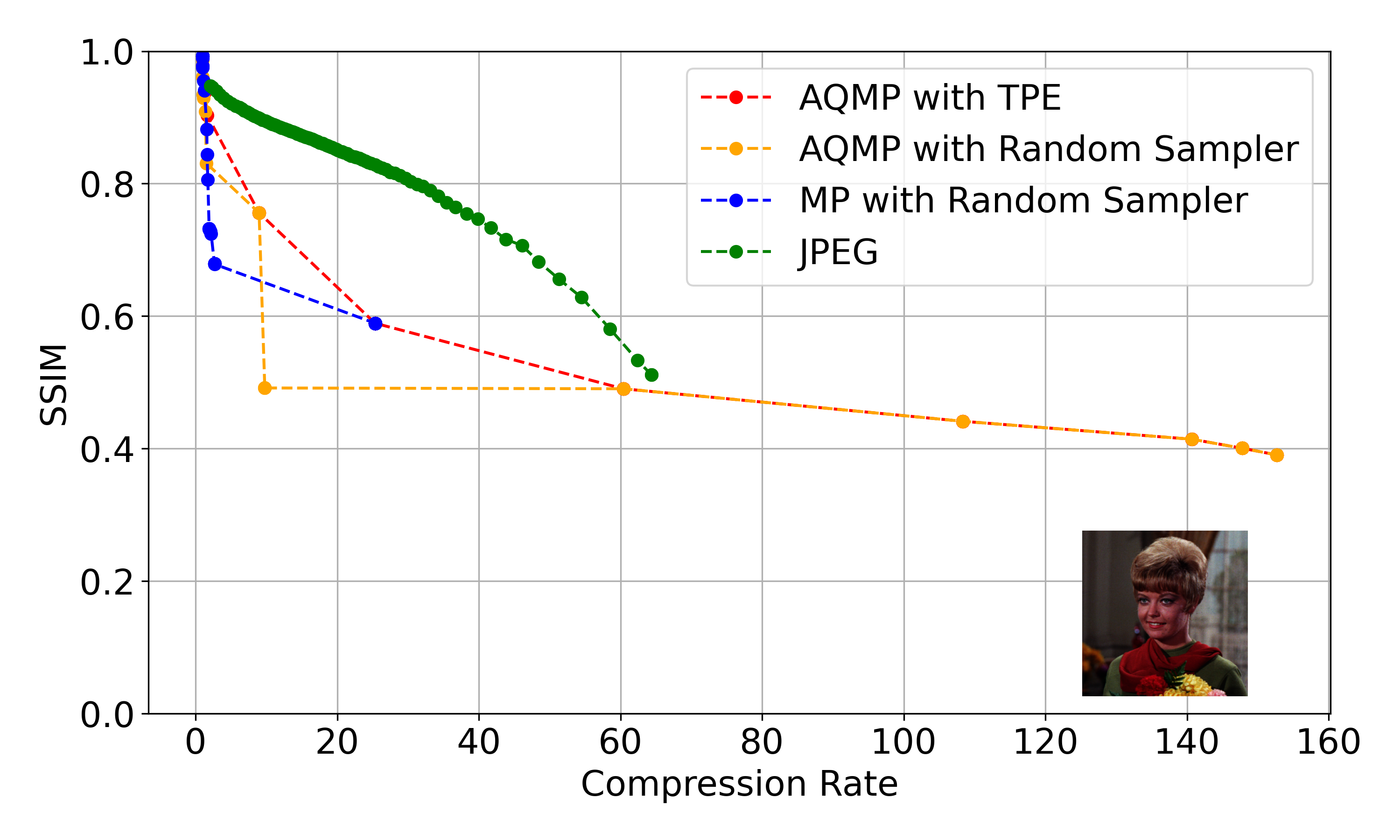}
    \caption{Comparison of AQMP with TPE, AQMP with Random Sampler, MP without quadtree refinement, and JPEG for the Female image.}
    \label{fig:comparison_female}
\end{figure}

\begin{figure}[h!]
    \centering
    \includegraphics[width=0.8\linewidth]{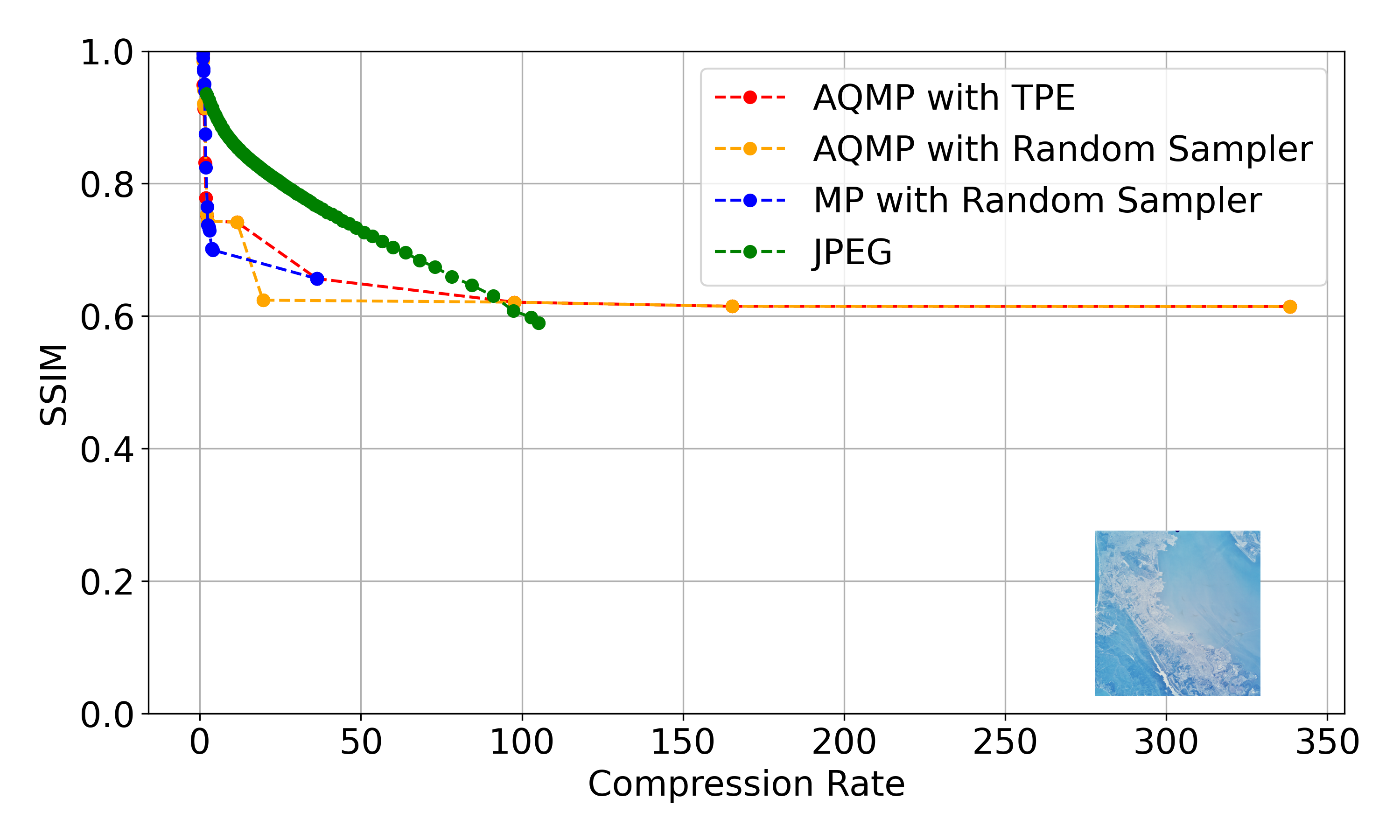}
    \caption{Comparison of AQMP with TPE, AQMP with Random Sampler, MP without quadtree refinement, and JPEG for the San Francisco image.}
    \label{fig:comparison_san_francisco}
\end{figure}

\begin{figure}[h!]
    \centering
    \includegraphics[width=0.8\linewidth]{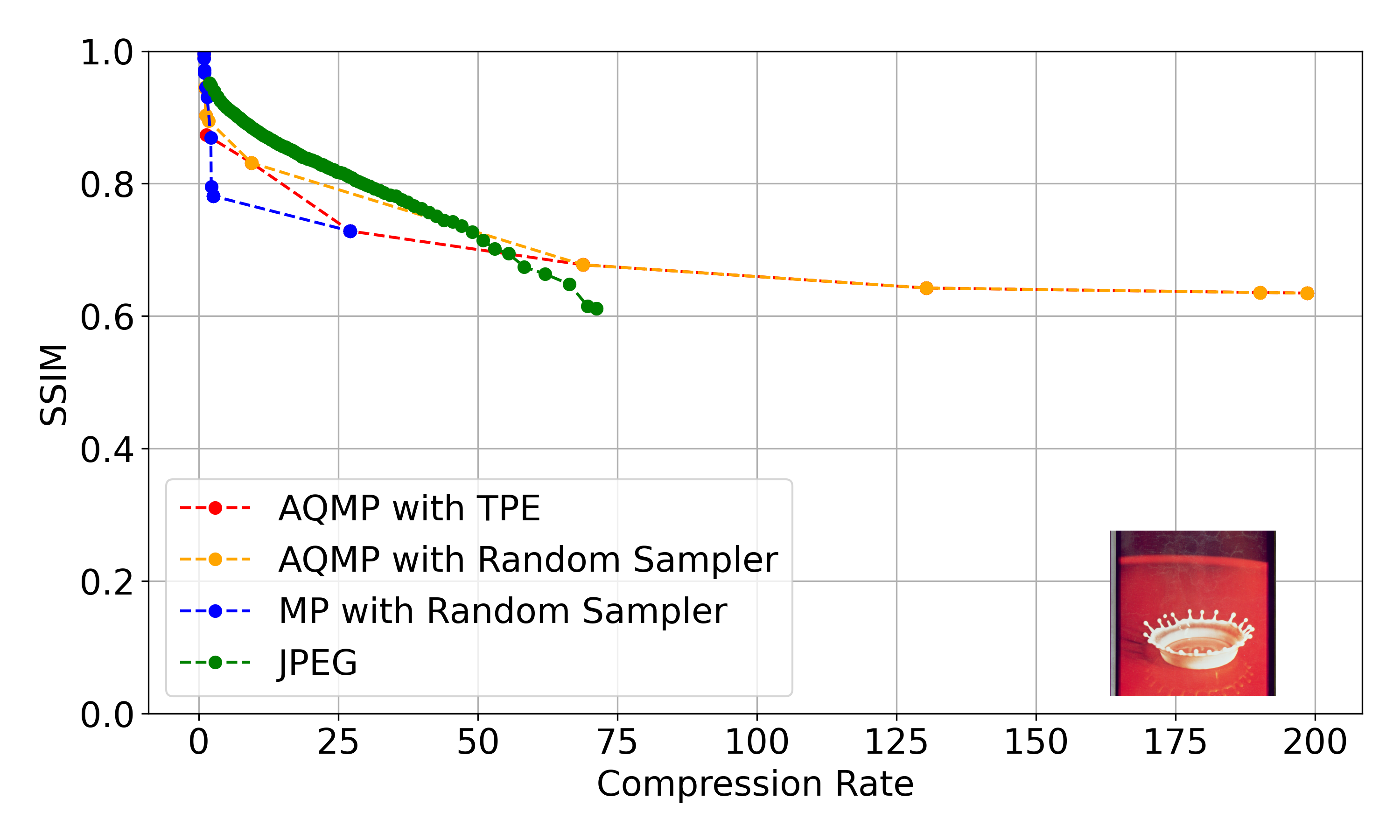}
    \caption{Comparison of AQMP with TPE, AQMP with Random Sampler, MP without quadtree refinement, and JPEG for the Splash image.}
    \label{fig:comparison_splash}
\end{figure}

\begin{figure}[h!]
    \centering
    \includegraphics[width=0.8\linewidth]{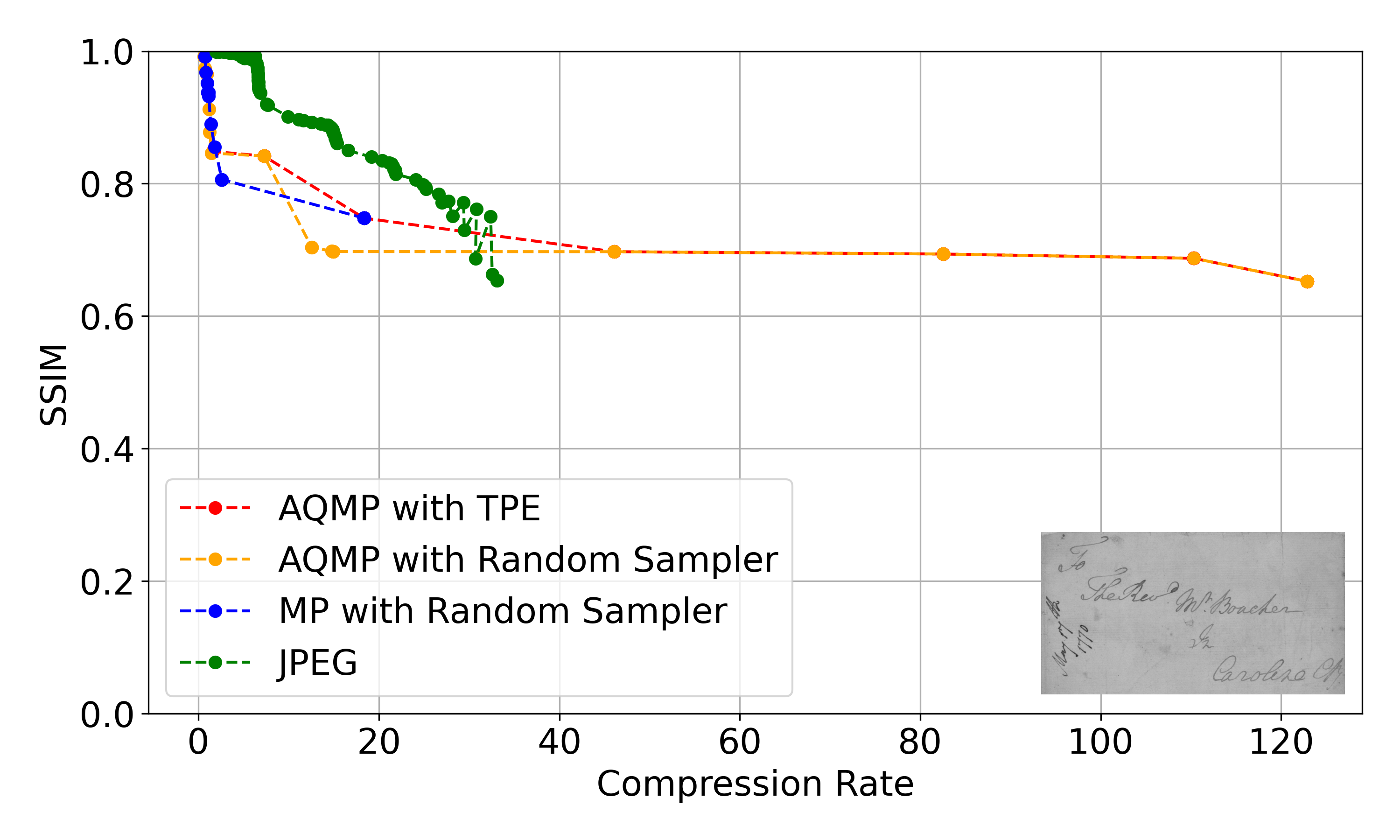}
    \caption{Comparison of AQMP with TPE, AQMP with Random Sampler, MP without quadtree refinement, and JPEG for the handwritten image `H02' of the International Document Image Binarization Contest (DIBCO) from the year 2010 \cite{Pratikakis2010}.}
    \label{fig:h02}
\end{figure}

\begin{figure}[h!]
    \centering
    \includegraphics[width=0.8\linewidth]{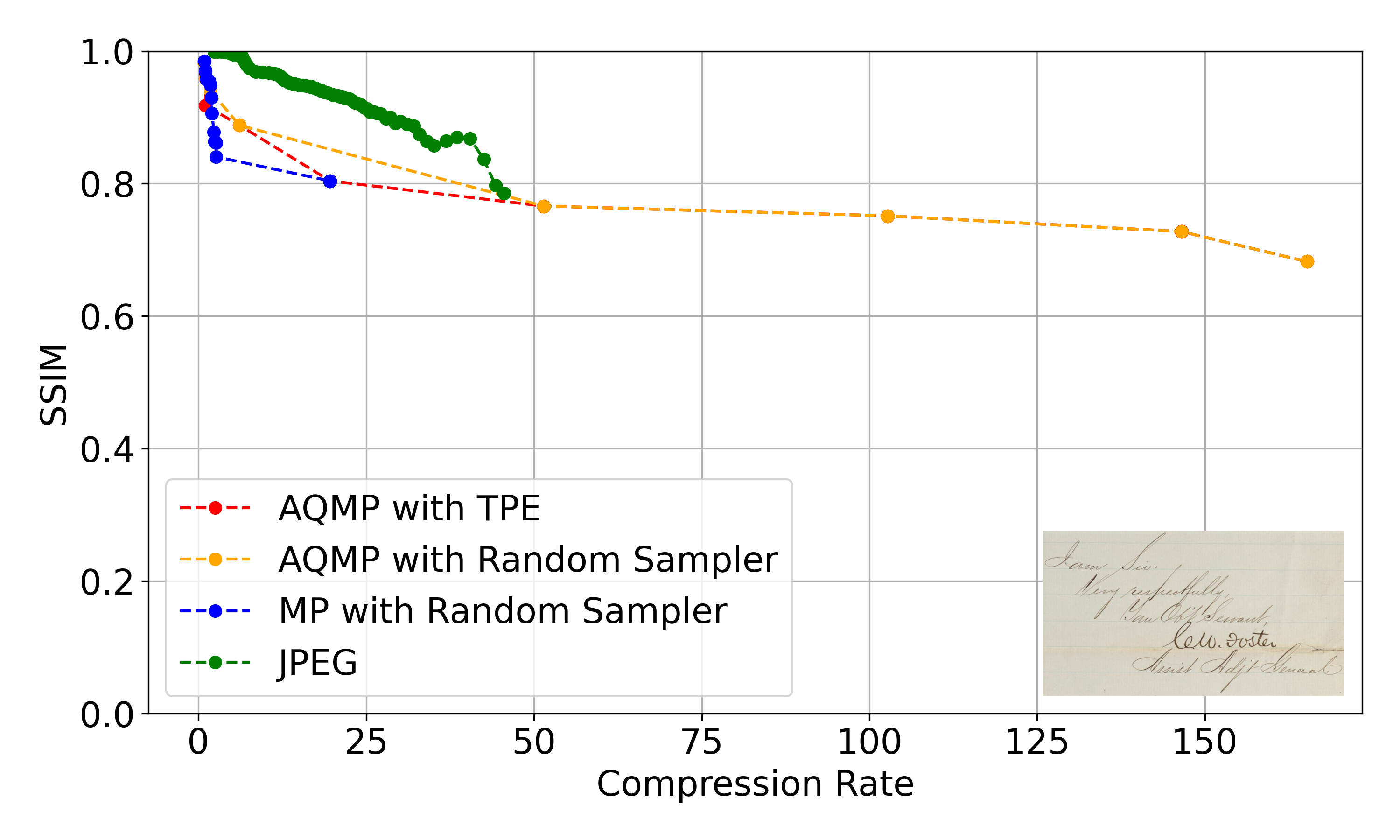}
    \caption{Comparison of AQMP with TPE, AQMP with Random Sampler, MP without quadtree refinement, and JPEG for the handwritten image `H09' of the International Document Image Binarization Contest (DIBCO) from the year 2010 \cite{Pratikakis2010}.}
    \label{fig:h09}
\end{figure}

One one hand, the comparison is based on test images of varying resolutions from the dataset \cite{sipi}: Female (256x256), Splash (512x512), and San Francisco (1024x1024). The results are presented as SSIM vs. Compression Rate curves, as shown in Figures [\ref{fig:comparison_female}], [\ref{fig:comparison_san_francisco}], and [\ref{fig:comparison_splash}]. On the other hand, comparisons are done with handwritten images of the dataset of the International Document Image Binarization Contest (DIBCO)\cite{Pratikakis2010}, as shown in Figures [\ref{fig:h02}] and [\ref{fig:h09}].

Overall we see that MP-based techniques, including AQMP, achieve higher compression ratios compared to JPEG. This is due to the ability of MP to adaptively represent image content using sparse representations, which is particularly effective for images with complex local structures. For low compression ratios, the JPEG algorithm performs well in preserving image quality. However, as the compression ratio increases, there is a point where JPEG becomes less efficient than AQMP techniques. This transition highlights the strength of AQMP in preserving quality at higher compression ratios. At low compression ratios, AQMP with TPE consistently outperforms both MP without quadtree refinement and AQMP with a random sampler. This demonstrates the effectiveness of TPE in optimizing hyperparameters to balance compression and quality. The random sampler, while still leveraging the quadtree structure, is less efficient in finding optimal configurations, leading to lower SSIM values at some specific compression ratios. This will clearly depends on how similar are the random parameters to the efficient ones. 

The results demonstrate that AQMP with TPE is more effective than MP and AQMP with random sampling for achieving high compression ratios while preserving image quality. It outperforms JPEG at higher compression ratios and provides better quality preservation compared to MP without quadtree refinement and AQMP with a random sampler. These findings highlight the importance of both the quadtree refinement and the use of TPE for hyperparameter optimization in achieving optimal performance.

\subsection{JPEG vs AQMP: compression-rate vs SSIM} \label{jpeg-vs-aqmp}

We complement the comparisons in Sec.~\ref{AQMP-comparison} with an aggregated view contrasting JPEG against AQMP using compression rate (CR) versus SSIM curves across three representative datasets: USC-SIPI, DIBCO, and DIV2K. For each dataset, we sweep AQMP hyperparameters and JPEG quality factors to trace the operating curves, and we report the best observed trade-offs for each method. In all plots below, the horizontal axis is the compression rate, CR $= \text{size}_{\text{original}}/\text{size}_{\text{compressed}}$ (higher is better), and the vertical axis is SSIM (higher is better).

Qualitatively, we observe a consistent pattern already suggested by Sec.~\ref{AQMP-comparison}:
\begin{itemize}
	\item At low compression (small CR), JPEG remains very strong and often slightly ahead in SSIM due to its mature quantization and deblocking pipeline.
	\item As compression becomes aggressive (larger CR), AQMP retains perceptual quality better, pushing the frontier toward higher CR at the same SSIM.
	\item The crossover depends on the content: highly textured or structured images tend to favor AQMP earlier, whereas smooth photographic content can favor JPEG longer.
\end{itemize}

\begin{figure*}[h!]
\centering
	 \makebox[\textwidth][c]{%
	\begin{subfigure}[t]{0.6\textwidth}
		\centering
		\includegraphics[width=\linewidth]{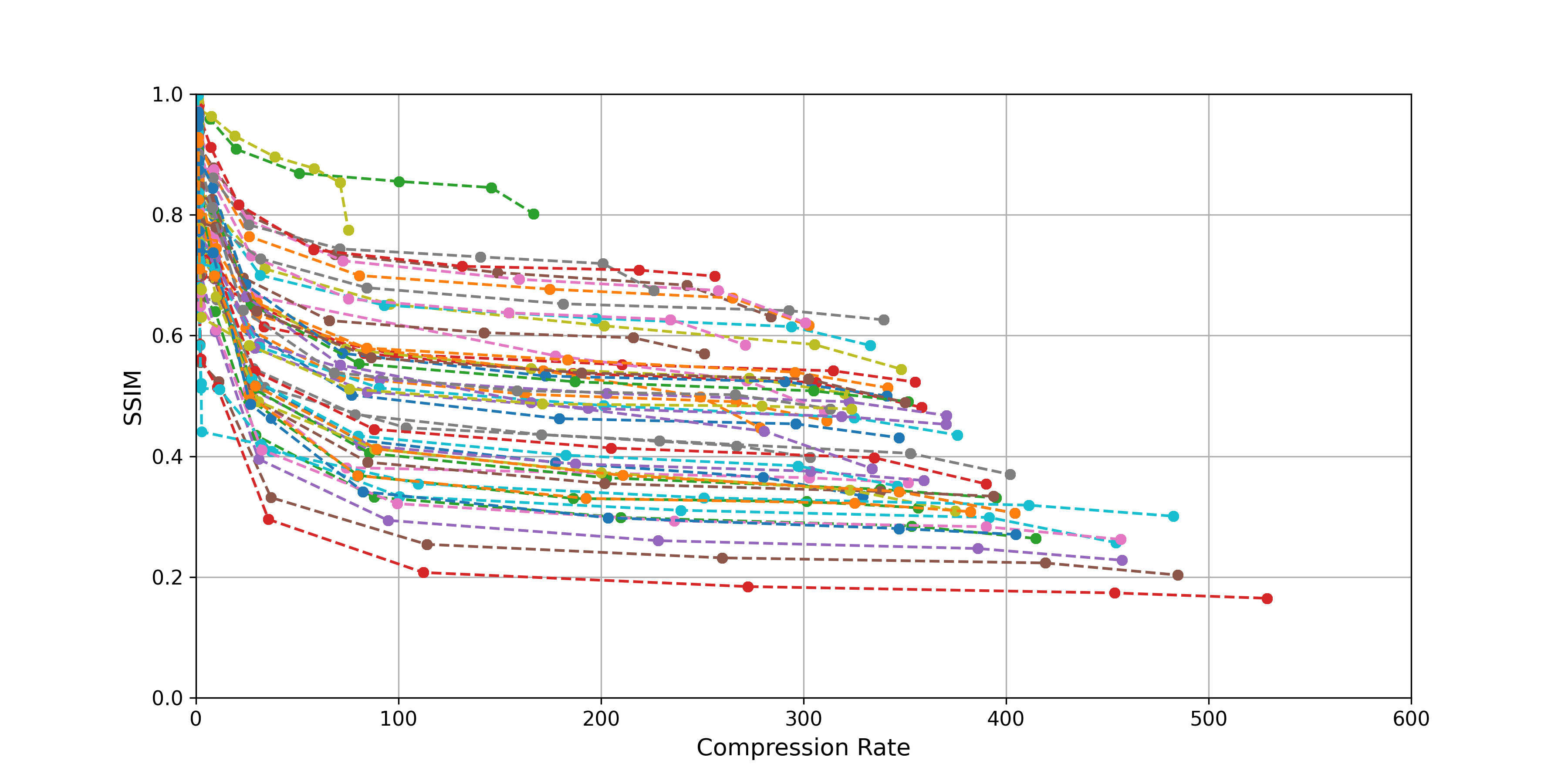}
		\caption{DIV2K — AQMP}
	\end{subfigure}
	\hfill
	\begin{subfigure}[t]{0.52\textwidth}
		\centering
		\includegraphics[width=\linewidth]{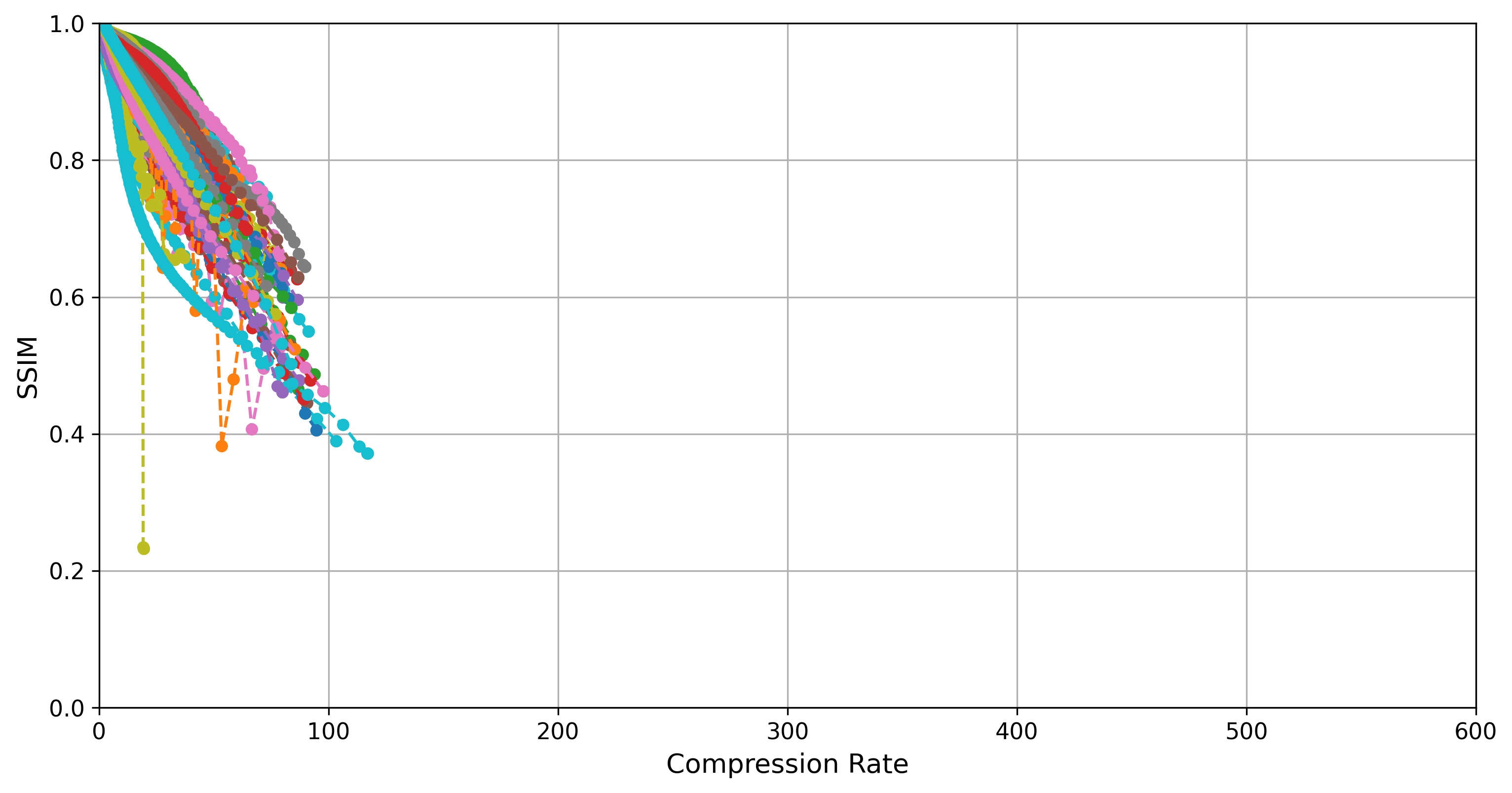}
		\caption{DIV2K — JPEG}
	\end{subfigure}
	 }
	\caption{Compression-rate vs SSIM curves for AQMP (left) and JPEG (right) across images of DIV2K dataset. AQMP curves are obtained by sweeping hyperparameters along the Pareto front; JPEG curves by varying quality factors.}
	\label{fig:jpeg_vs_aqmp_curves}
\end{figure*}

Figure~\ref{fig:jpeg_vs_aqmp_curves} shows CR–SSIM curves for the three datasets. Each row corresponds to a dataset; left column shows AQMP results, right column shows JPEG results. These curves were generated from the same image sets and settings used throughout this paper; details on metrics appear in Sec.~\ref{results}.

\begin{figure*}[h!]
    \centering
    \makebox[\textwidth][c]{%
        \begin{subfigure}[t]{0.6\textwidth}
            \centering
            \includegraphics[width=\linewidth]{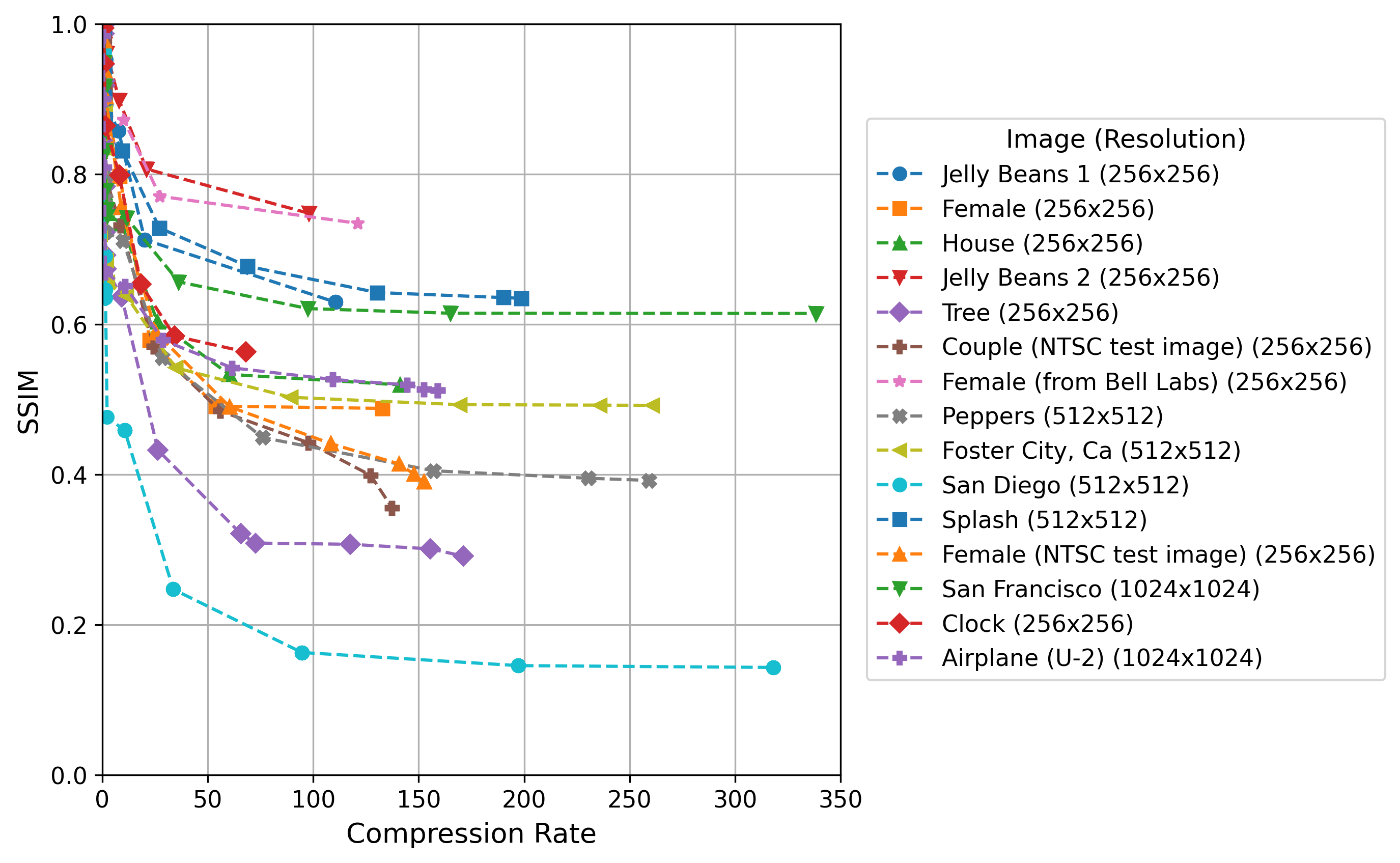}
            \caption{USC-SIPI — AQMP}
        \end{subfigure}
        \hfill
        \begin{subfigure}[t]{0.6\textwidth}
            \centering
            \includegraphics[width=\linewidth]{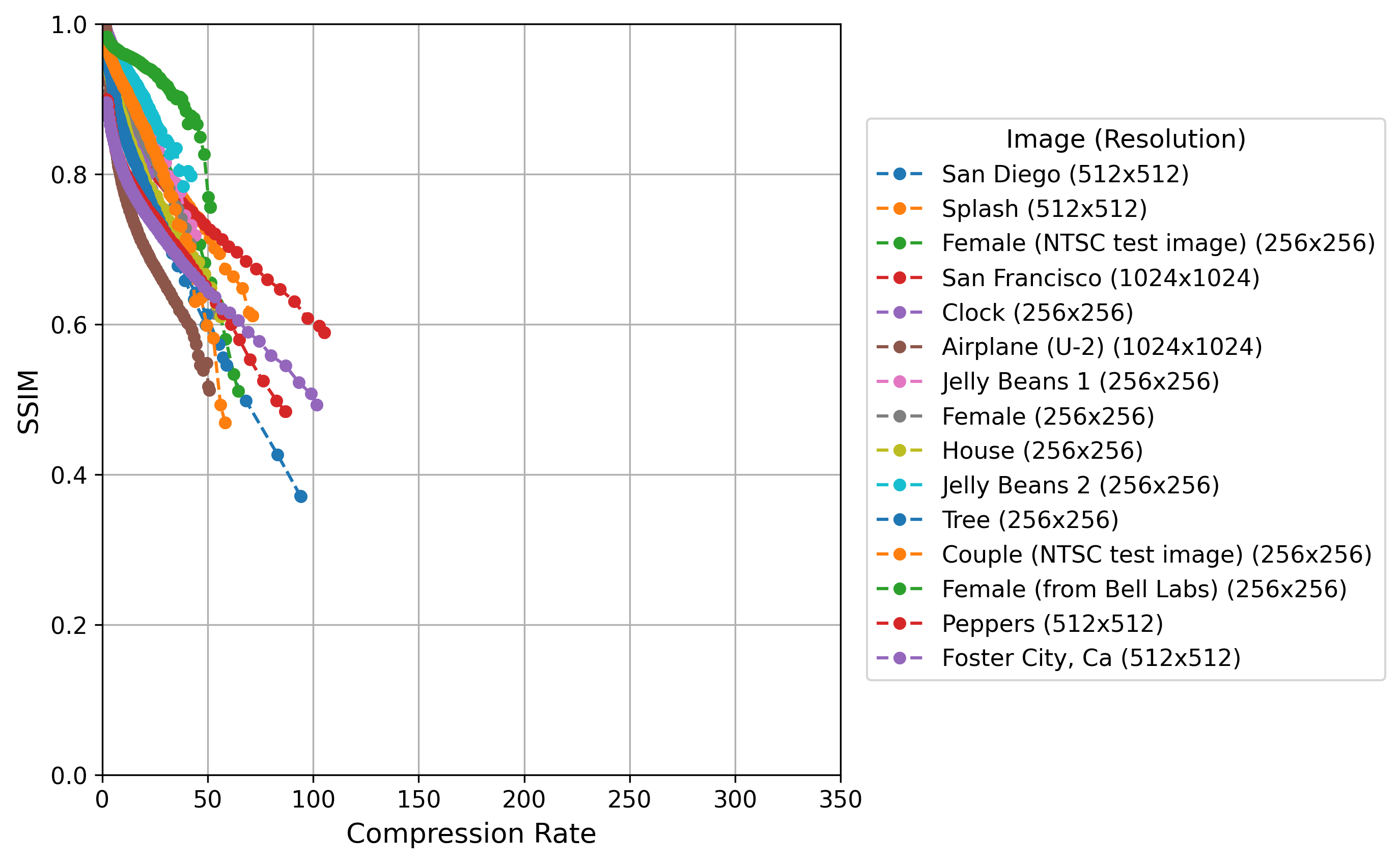}
            \caption{USC-SIPI — JPEG}
        \end{subfigure}
    }
    \caption{Compression-rate vs SSIM curves for AQMP (left) and JPEG (right) across images of the USC-SIPI dataset. AQMP curves are obtained by sweeping hyperparameters along the Pareto front; JPEG curves by varying quality factors.}
\end{figure*}

	\begin{figure*}[h!]
		\centering
	   \makebox[\textwidth][c]{%
	\begin{subfigure}[t]{0.6\textwidth}
		\centering
		\includegraphics[width=\linewidth]{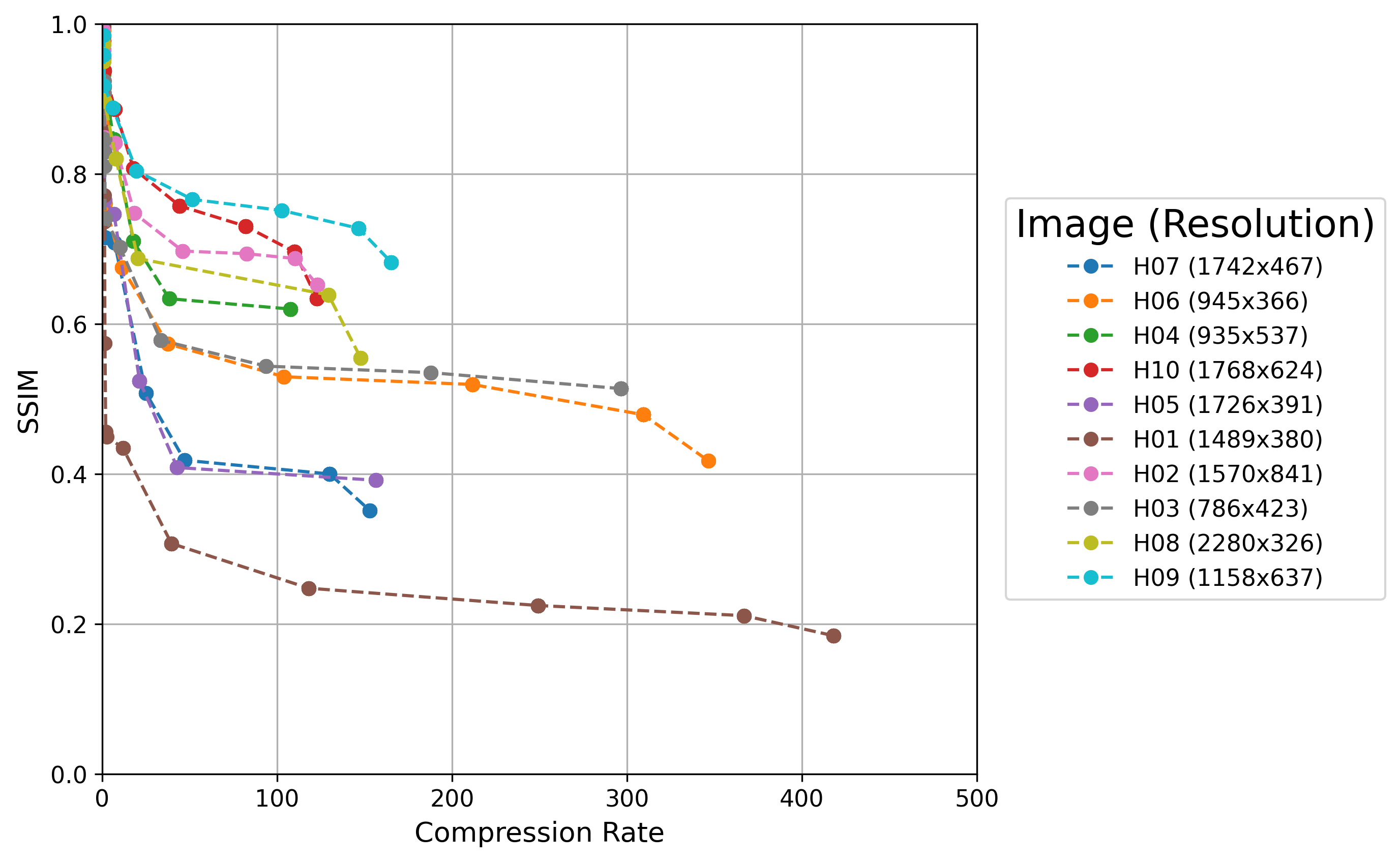}
		\caption{DIBCO — AQMP}
	\end{subfigure}
	\hfill
	\begin{subfigure}[t]{0.6\textwidth}
		\centering
		\includegraphics[width=\linewidth]{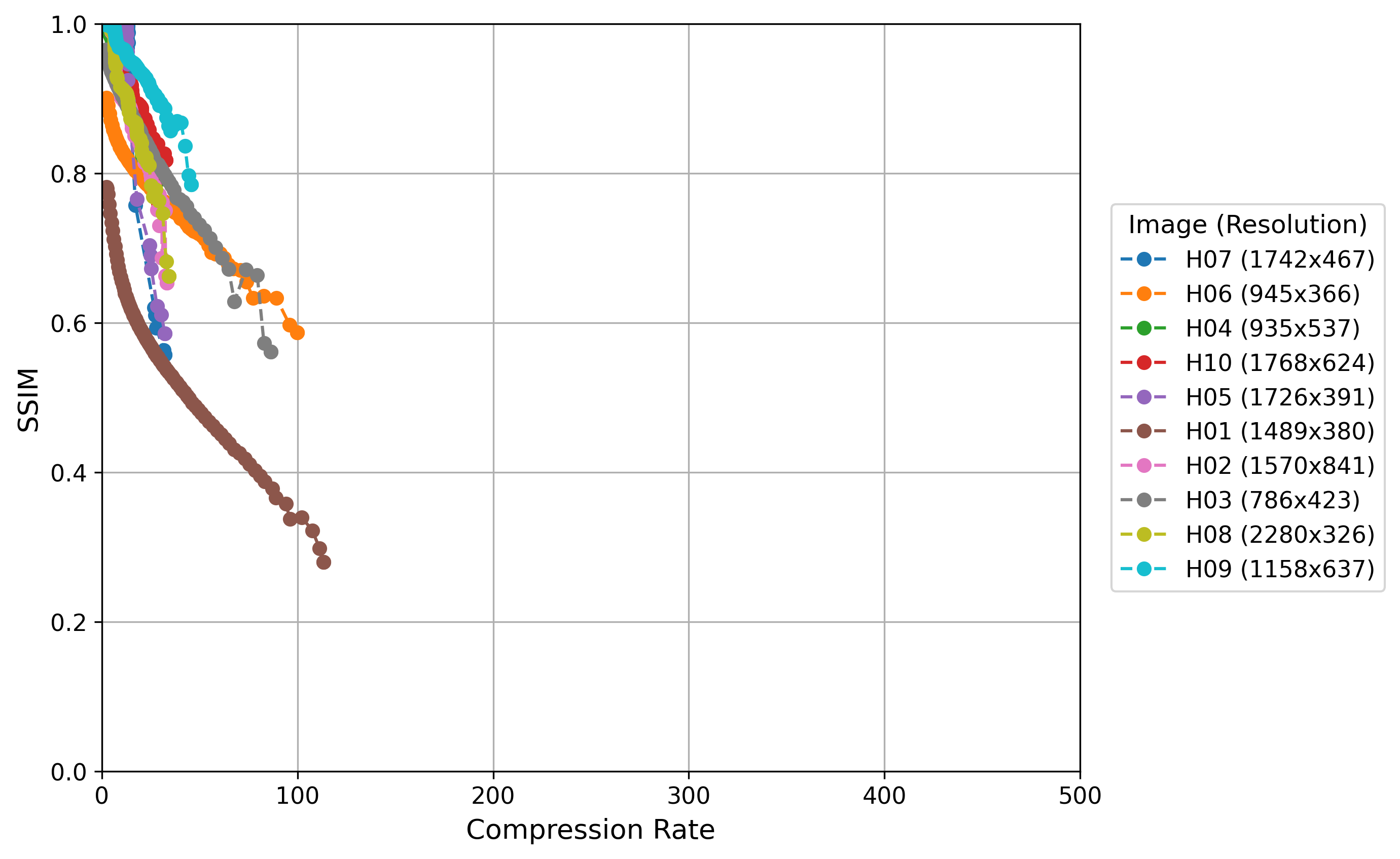}
		\caption{DIBCO — JPEG}
	\end{subfigure}
	   }
	\caption{Compression-rate vs SSIM curves for AQMP (left) and JPEG (right) across images of DIBCO dataset. AQMP curves are obtained by sweeping hyperparameters along the Pareto front; JPEG curves by varying quality factors.}
\end{figure*}

Key takeaways from Fig.~\ref{fig:jpeg_vs_aqmp_curves}:
\begin{itemize}
	\item USC-SIPI: AQMP begins to dominate beyond medium compression, offering higher CR at similar SSIM, aligned with the results in Figs.~\ref{fig:comparison_female}–\ref{fig:comparison_splash}.
	\item DIBCO: AQMP maintains structure in thin strokes and binary-like regions, sustaining SSIM at higher CR where JPEG degrades sharply.
	\item DIV2K: On natural photos, JPEG leads at very low CR, but AQMP narrows the gap and surpasses it at higher CR.
\end{itemize}

Overall, AQMP provides a favorable quality–compression trade-off in the high-compression regime, while JPEG remains hard to beat at very low compression. Depending on use case, one may prefer AQMP for bandwidth/storage constrained settings, or JPEG for archival-quality at modest CR.

\section{Discussions and Potential Extensions}\label{discussion}

In this work, we introduced a quadtree-based compression algorithm using Matching Pursuit capable of handling images at any resolution and
adapting the partition size to local image complexities. By autonomously and efficiently tuning a comprehensive hyperparameter set, the algorithm can flexibly balance compression ratio and image quality. Through sequential model-based optimization (cf. Fig.[\ref{fig:compression_results}]), we demonstrated that high levels of compression can be achieved while preserving visual fidelity.  

Beyond the per-image analyses, we also carried out a thorough, side-by-side comparison between JPEG and AQMP via compression-rate vs SSIM curves across multiple datasets (cf. Sec.\,\ref{jpeg-vs-aqmp}). This makes explicit the operating regimes where each codec is preferable. For high-resolution content (DIV2K, \textasciitilde2K images), the trends persist: JPEG leads at very low compression, while AQMP sustains higher SSIM at large compression rates. At these resolutions, container metadata overhead is negligible relative to content, so CR reflects compression more faithfully and the crossover toward AQMP can occur at slightly lower SSIM loss than in 256–512 px cases. We also observed AQMP’s quadtree+MP scales well to 2K resolutions, with the main trade-off being longer encode times (still practical in our experiments), while decoding remains lightweight.
When compressing the images of the datasets \cite{sipi, Pratikakis2010}, we found that the best compression rates of AQMP+TPE compared to JPEG are found for SSIM up to approximate values of  0.7 and 0.8. In these cases, we found up to a 4x compression rate for similar SSIM values, as shown in Section [\ref{AQMP-comparison}]. We demonstrated the advantages of AQMP + TPE, highlighting its ability to preserve image quality at higher compression ratios while maintaining greater fidelity to the original image at low compression rates compared to alternative approaches (e.g., AQMP + Random Sampler and Only MP). Our analysis revealed a consistent trend: JPEG remains the superior algorithm for low compression ratios, achieving unmatched quality in this regime. However, JPEG exhibits a rapid decline in quality at higher compression ratios, limiting its practicality for aggressive compression scenarios. In contrast, AQMP + TPE enables improved quality retention at low compression ratios compared to its alternatives, though it does not yet match JPEG’s performance in this range.
These findings suggest that future work could focus on refining the AQMP framework—for instance, by optimizing its parameterization or sampling strategies—to bridge the performance gap with JPEG at low compression ratios while retaining its robustness in high-compression regimes.

On the other hand, the current design relies on numerous hyperparameters (\texttt{min\_sparsity}, \texttt{min\_n}, \texttt{max\_n}, \texttt{max\_error} and \texttt{a\_cols}), making it challenging to determine the best configuration for achieving either higher quality or smaller file size. However, this is not a practical impediment; for instance, one can use higher values of \texttt{max\_n} and \texttt{a\_cols} to achieve higher compression rates, or high and low values in \texttt{max\_error} and \texttt{min\_sparsity} to achieve higher SSIM values, respectively. A future direction of this work might be in   reducing these into a smaller, more manageable subset (ideally just \texttt{max\_n} and \texttt{max\_error}), simplifying and speeding up the HPO optimization. 

Possible applications of AQMP, where a good compression ratio is needed but not necessarily a high SSIM, include limited-bandwidth scenarios and environments with low storage capacity. In the future, the AQMP algorithm might be extended to support video compression by leveraging temporal redundancy and motion estimation. This can be done by incorporating techniques such as block matching or optical flow to identify and exploit similarities between consecutive frames, while each frame can be compressed by using our AQMP method. 
Future work could also explore the integration of machine learning models to further enhance compression efficiency and adaptiveness to different types of video content.

The code developed for this project is open source and free available under license MIT through GitHub at \cite{aqmp-github}. 

\section{Acknowledgements}

This research has been supported by CONICET, Argentina. We thank Alfredo Nava-Tudela and Andrés Pagliano for initial discussions a while ago on some aspects of the approach followed in this paper. 

\appendix

\section{Workflow Summary} \label{code_workflow}
\subsection{Encoder}

Given an input image and an error tolerance $\varepsilon$, to encode the 
image apply the following steps:

\begin{enumerate}
\item Pre-processing step: Separate color channels using format $\mathrm{YC_BC_R}$
\item MP step: Apply the $mp\_coder$ algorithm as defined in section 
	\ref{description-encoder} using the following parameters for illustration purposes:
	\begin{enumerate}
	\item $M = 128$
	\item $S = \{8, 16, 32 \}$
	\item $\mathbf{D}_N = \textup{DCT}_{N^2\times M} \;|\; \textup{Haar}_{N^2\times M}$ for each $N \in S$ 
	\item $unravel(\mathbf{A}) =$ concatenate all columns of $\mathbf{A}$
	\item $min\_sparcity(\varepsilon, N) = \left \{
	\begin{matrix}
       128 & \quad\text{if}\ N = 8 \\
       2   & \quad\text{if}\ N = 16 \\
       3   & \quad\text{if}\ N = 32
     \end{matrix} \right.  $ 
	\end{enumerate}
\item Lossless coding step: Apply DEFLATE coding \cite{deutsch96} to the result of the MP compressor
\end{enumerate}

$\textup{DCT}_{N^2\times M}$ and $\textup{Haar}_{N^2\times M}$ are the overcomplete dictionaries 
of size $N^2\times M$ using Discrete Cosine Transform and Haar Wavelets, see for example \cite{tudela12}.

\subsection{Decoder}

Given a compressed file $f$, to reconstruct the image apply the following steps: 

\begin{enumerate}
\item Lossless decoding: Apply the DEFLATE decoding \cite{deutsch96} process to $f$.
\item MP decoding step: For each channel, apply the $mp\_decoder$ algorithm as defined in 
	section \ref{description-decoder} using the following parameters:
	\begin{enumerate}
	\item $M = 128$
	\item $S = \{8, 16, 32 \}$
	\item $\mathbf{D}_N = \textup{DCT}_{N^2\times M} \;|\; \textup{Haar}_{N^2\times M}$ for each $N \in S$ 
	\item $unravel(\mathbf{A}) =$ concatenate all columns of $\mathbf{A}$
	\end{enumerate}
\item Post-processing: Convert channels to RGB format (Optional). 
\end{enumerate}

\end{document}